\pdfoutput=1
\documentclass{preprint}

\usepackage{amsmath,amsfonts,bm}

\def\eqref#1{equation~\ref{#1}}

\def\1{\bm{1}}

\DeclareMathAlphabet{\mathsfit}{\encodingdefault}{\sfdefault}{m}{sl}
\SetMathAlphabet{\mathsfit}{bold}{\encodingdefault}{\sfdefault}{bx}{n}

\usepackage[utf8]{inputenc}
\usepackage[T1]{fontenc}
\usepackage[english]{babel}
\usepackage{microtype}

\usepackage{amsmath}
\usepackage{amssymb}
\usepackage{amsfonts}
\usepackage{amsthm}
\usepackage{bm}
\usepackage{nicefrac}

\usepackage{booktabs}
\usepackage{array}
\usepackage{multirow}
\usepackage{tabularx}
\usepackage{colortbl}

\usepackage{graphicx}
\usepackage{float}
\usepackage{wrapfig}
\usepackage{caption}

\newcommand{\safeincludegraphics}[2][]{%
  \IfFileExists{#2}{%
    \includegraphics[#1]{#2}%
  }{%
    \fbox{\parbox[c][0.3\textwidth][c]{0.9\textwidth}{\centering Missing figure: \texttt{\detokenize{#2}}}}%
  }%
}

\usepackage{algorithm}
\usepackage{algorithmicx}
\usepackage[noend]{algpseudocode}
\floatname{algorithm}{Algorithm}

\usepackage{enumitem}
\usepackage{xspace}
\usepackage{verbatim}
\usepackage{balance}
\usepackage[normalem]{ulem}
\useunder{\uline}{\ul}{}

\usepackage{xcolor}
\usepackage{amssymb}
\definecolor{failred}{HTML}{9B3038}

\usepackage{hyperref}
\usepackage{url}
\hypersetup{
    colorlinks=true,
    linkcolor=linkblue,
    urlcolor=linkblue,
    citecolor=linkblue,
}

\newcommand{\ie}{\emph{i.e., }}
\newcommand{\eg}{\emph{e.g., }}

\title{Multi-Agent Harness Optimization Across Node, Edge, and Graph}

\author[1,*]{Xinyu Lin}
\author[2,*]{Yufan Liu}
\author[3,\dagger]{Chen Xu}
\author[2]{Denghong Luan}
\author[4]{Yongqi Li}
\author[2]{Fuli Feng}
\author[2,\dagger]{Wenjie Wang}

\affiliation{$^1$National University of Singapore \quad $^2$University of Science and Technology of China \\ $^3$Carnegie Mellon University \quad $^4$The Hong Kong Polytechnic University}
\contribution{$^*$Equal contribution \quad $^\dagger$Corresponding authors}

\abstract{Multi-agent systems (MAS) show strong potential for complex task solving, but realizing this potential requires effective agent harnesses, which impose infrastructural constraints on execution and collaboration.
Existing approaches rely on either manual design, which is labor-intensive and task-dependent, or automatic generation, which can overlook inter-agent and group-level failures, while emphasizing the individual agent harness. Effective MAS harnesses must therefore cover individual execution (\ie node), inter-agent collaboration (\ie edge), and group coordination (\ie graph).
However, directly extending the automatic optimization loop of failure diagnosis and constraint generation across these levels introduces two challenges: failure diagnosis duplication, where one underlying failure is repeatedly diagnosed through its propagated symptoms, and constraint candidate explosion, where numerous proposed repairs incur excessive execution costs. We propose MASH, a framework for efficient MAS harness optimization across all three levels. MASH performs structured failure diagnosis by tracing observed symptoms to shared gaps between the information a decision requires and receives, reducing redundant diagnoses and repeated repairs. It then generates coordinated cross-level constraints grounded in each diagnosed failure and applies progressively more expensive quality checks to filter weak candidates before they enter the next optimization round. Experiments across four benchmarks and four MAS topologies demonstrate that MASH consistently outperforms the evaluated baselines, supporting the effectiveness of harness optimization tailored to multi-agent execution. The code is available  \href{https://github.com/GeniusNew/MASH_code_public}{here}.}

\date{\today}
\correspondence{Chen Xu, Wenjie Wang}
\metadata[Code]{\url{https://github.com/GeniusNew/MASH_code_public}}
\input{sections/mash_visual_preamble.tex}
\begin{document}

\maketitle

\section{Introduction}

Multi-agent systems (MAS) achieve strong task-solving capability by dividing work across specialized agents and executing in parallel, showcasing remarkable performance on tasks such as programming~\citep{hong2024metagpt,qian-etal-2024-chatdev}, mathematical reasoning~\citep{pmlr-v235-du24e,chen-etal-2024-reconcile}, and deep research~\citep{hadfield2025research,camara2026selfoptimizingmultiagentsystemsdeep}.
However, their performance depends not only on the capabilities of individual models, but also on the \textbf{\textit{agent harness}}, \ie the \textit{infrastructural constraints} governing tool use, information exchange, and execution control~\citep{wu2023autogenenablingnextgenllm}.
Inadequate constraints can prevent otherwise capable agents from working effectively together, leading to failures in execution, collaboration, and verification~\citep{cemri2025multi, ning2026codeagentharness}.
Therefore, it is essential to build an effective MAS harness, unleashing the full potential of MAS and facilitating real-world applications.

\begin{figure}[t]
    \centering
\includegraphics[width=\linewidth]{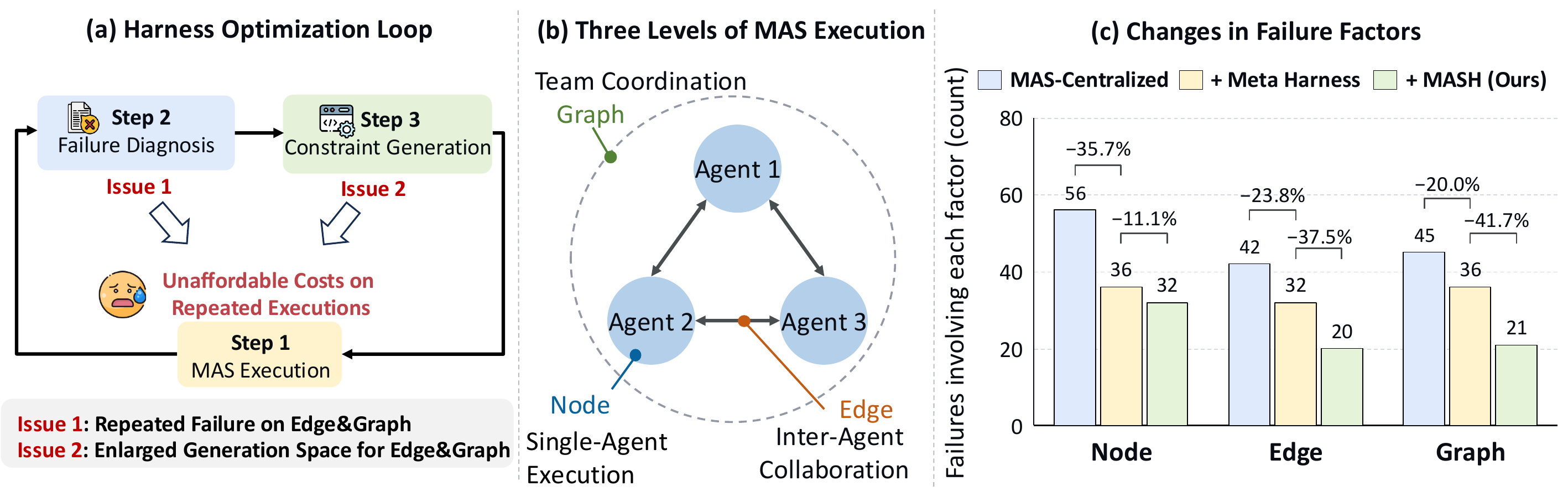}
    \caption{Motivation for MAS harness optimization. (a) Extending optimization to MAS introduces redundant diagnosis and more constraint candidates, increasing execution costs. (b) MAS execution spans node, edge, and graph levels with inter-agent information exchange. (c) On Finance-Agent, existing harness optimization is less effective at addressing edge- and graph-level failures than node-level failures. }
    \label{fig:intro-motivation}
\end{figure}

Recent research typically achieves harness (\ie infrastructural constraints) construction through an iterative optimization loop~\citep{lee2026metaharness,lin2026ahe,chen2026harnessx,zhang2026selfharness} as shown in Figure~\ref{fig:intro-motivation}(a).
Specifically, given an agent task (\eg deep research), the optimization loop typically contains three steps:
1) \textit{task execution}, which runs agents under each proposed infrastructural constraint and collects trajectories\footnote{In the initial optimization round, agents execute under the original harness.}.
Based on execution results, useful constraints are incorporated into the harness.
2) \textit{failure diagnosis}, which examines the collected trajectories to diagnose infrastructural failures (\eg evidence lost during agent handoffs) and determine whether the proposed constraints address them;
3) \textit{constraint generation}, which proposes new constraints based on the diagnosed failures.
With the generated constraints, it then goes back to the first step (\ie task execution) for next-round optimization.

Despite their effectiveness, existing automatic harness-construction methods fail to generalize well to MAS, yielding only partial harnesses.
Specifically, existing methods primarily diagnose failures and generate constraints around an individual agent's interaction with its environment.
Yet, MAS execution involves three levels, including single-agent execution (\ie node), inter-agent collaboration (\ie edge), and agent group coordination (\ie graph).
Directly applying existing methods can over-emphasize the constraints on single-agent execution, thus leaving failures in edge and graph levels unfixed (see Figure~\ref{fig:intro-motivation} for empirical evidence).
Therefore, it is essential to build infrastructural constraints for MAS explicitly across the node, edge, and graph levels.

A natural solution is to explicitly diagnose failures  across the three levels and then propose level-specific constraints.
However, in MAS, interactions among agents allow a node-level failure to propagate to the edge and graph levels.
Diagnosing these failures can produce repeated diagnoses and similar constraint candidates that target the same underlying issue.
The key is therefore to retain a small set of high-quality candidates while eliminating redundant or ineffective ones.
To achieve this, we posit two principles for failure diagnosis and constraint generation steps, respectively:
1) \textbf{Failure deduplication}, which aims to reduce repeated diagnoses that may generate similar constraints.
Repeated diagnoses arising from failure propagation can yield similar constraint candidates, each requiring a separate MAS execution to collect trajectories for the next round.
This wastes computation on the same underlying issue.
It is therefore important to merge diagnoses that share a root cause, retaining one diagnosis per distinct cause.
2) \textbf{{Constraint candidate filtering}}, which aims to avoid expensive MAS executions on low-quality candidates.
Generating constraints across three levels can produce a large number of constraint candidates, each requiring a separate MAS execution for next-round diagnosis.
Passing all candidates into subsequent rounds therefore causes overwhelming computational burdens.
To address this issue, the key is to quickly filter out low-quality candidates before candidates enter the next round, reserving execution budget for promising constraints.

Following these principles, we propose \textbf{MASH} (\textbf{M}ulti-\textbf{A}gent \textbf{S}ystem \textbf{H}arness optimization), an efficient MAS harness optimization framework across node, edge, and graph.
At the \textit{failure diagnosis step}, we introduce a structured failure diagnosis strategy, which aims to trace the root cause of failures and reduce repeated diagnosis.
It organizes each diagnosis as a structured tuple indicating what required information is missing and where the information gap first arises during execution (detailed descriptions are in Section~\ref{sec:diag}).
By grouping structured diagnoses that concern the same required information, MASH reduces repeated diagnoses and avoids generating similar constraints.
For \textit{the constraint generation step}, MASH first generates constraints explicitly for node, edge, and graph levels to cover the constraint space of MAS.
To prevent low-quality candidates from entering the next-round execution, we introduce a cascaded filtering strategy consisting of checks on syntax and semantics (\ie whether the generated constraint is executable and consistent with its proposed changes), effectiveness (\ie whether the constraint is useful in the specific case), and robustness (\ie whether the constraint generalizes across diverse tasks), respectively.
The remaining candidates that pass the checks will then be executed and diagnosed for the next round.
We evaluate MASH on four datasets (\ie BrowseComp-Plus, WorkBench, Plancraft, Finance) across multiple workflows and LLM backbones, demonstrating its effectiveness across individual-agent execution, inter-agent interactions, and team coordination, with an average reduction in node, edge, and graph-level failures of 7.7\%, 25.5\%, and 29.1\%, respectively. Extensive analyses of constraint quality, optimization efficiency, and generalization further validate the effectiveness, efficiency, and generalizability of our proposed method.

The main contributions of this work are summarized as follows:
\begin{itemize}[leftmargin=1.4em,itemsep=2pt,topsep=2pt]
\item We motivate harness optimization across node, edge,
and graph levels in MAS and formulate two guiding principles,
failure deduplication and constraint candidate filtering,
to reduce redundant proposals and unnecessary MAS executions.

\item We propose MASH, an efficient MAS harness optimization
framework that combines structured failure diagnosis with
cross-level constraint generation and fast filtering to
retain high-quality constraints while reducing optimization costs.

\item We conduct extensive experiments on Finance-Agent,
BrowseComp-Plus, WorkBench, and Plancraft across multiple
workflows and backbone LLMs, demonstrating the effectiveness,
efficiency, and robustness of MASH.
\end{itemize}

\section{Related Work}

\subsection{LLM-based Multi-Agent Systems}
LLM-based MAS improve task solving through specialized roles, task decomposition, and inter-agent communication.
Frameworks such as MetaGPT and ChatDev organize agents through structured collaboration protocols~\citep{hong2024metagpt,qian-etal-2024-chatdev},  while subsequent work explores communication topologies and coordination strategies~\citep{zhou2025multi,li2026assemble}. Their benefits depend on task structure, model capability, and coordination overhead~\citep{kim2026capable}.
However, distributing execution also introduces opportunities for information loss, inter-agent misalignment, and incomplete verification~\citep{hadfield2025research,cemri2025multi}.
Empirical failure analyses identify system design and coordination deficiencies as important sources of unsuccessful execution, beyond limitations of individual models~\citep{zhang2025agent,zhang2025agentracer,kong2025aegis}, while TraceElephant emphasizes the importance of complete execution context for diagnosis~\citep{traceelephant2026}.
These observations motivate examining the infrastructure governing agent behavior and interaction alongside the underlying models.
Our work constructs a harness for MAS: infrastructural constraints that regulate individual execution, inter-agent information exchange, and group coordination to address failures across these interaction levels.

\subsection{Agent Harness Optimization}
Agent harnesses govern tool access, context management, and execution control~\citep{ning2026codeagentharness,lee2026metaharness,vero2026}.
Beyond manual design, recent methods use execution feedback to optimize prompts and context~\citep{agrawal2026gepa,zhang2025ace,kang2026acon}, memory and tools~\citep{zhang2026memevolve,zhang2026rewardharness}, and broader executable harness components~\citep{lee2026metaharness,lin2026ahe,xu2026lifeharness}.
Further work improves candidate generation and verification~\citep{chen2026harnessx,zhang2026selfharness,harnesslens2026,wu2026modularrsi}, or adapts harnesses across successive tasks and interactions~\citep{he2026evotest,continualharness2026,evoharness2026}.
We provide a detailed discussion in Appendix~\ref{app:harness-related-work}. In MAS, harness optimization needs to additionally address failures in inter-agent collaboration and group coordination.
Failure propagation across these levels can produce repeated diagnoses and similar constraint candidates, increasing evaluation costs.

\subsection{Multi-Agent Workflow Optimization and Evolution}
Prior work improves multi-agent systems through workflow search and architecture selection~\citep{zhang2024aflow,maas2025,wang2025evoagentx}, coordinated prompt optimization~\citep{hong2026maspob,wang2026maspo}, and the evolution of collaboration mechanisms and team organization~\citep{lee2026rhi,hao2026metateam}.
Data Canvas further uses provenance to trace errors and selectively replay affected workflow components~\citep{yi2026datacanvas}.
Building on these advances, MASH focuses on efficient harness optimization under fixed agents, communication topology, and execution budgets.
It links failures across levels to shared root causes and combines coordinated constraint generation with fast filtering, reducing repeated diagnosis and unnecessary candidate evaluation.

\section{Problem Formulation}
\label{sec:formulation}

\paragraph{MAS harness.}
An MAS $\mathcal M$ executes tasks under a harness $H$, a set of infrastructural constraints governing its execution.
MAS involve not only interactions between individual agents and their environments, but also information exchange among agents and coordination across the group.
We therefore define an MAS harness across three levels:
\[
H=(C^{\mathrm n},C^{\mathrm e},C^{\mathrm g}),
\]
where node constraints $C^{\mathrm n}$ govern individual agents' execution, edge constraints $C^{\mathrm e}$ govern inter-agent information exchange, and graph constraints $C^{\mathrm g}$ govern group coordination.

\paragraph{Harness optimization objective.}
Our objective is to find a harness that maximizes expected task performance:
\[
H^\star \in \arg\max_{H\in\mathcal H}
\mathbb E_{x\sim\mathcal D}
\left[s\bigl(\operatorname{Exec}(\mathcal M[H],x)\bigr)\right],
\]
where $\mathcal D$ is the task distribution, $\operatorname{Exec}$ returns an execution trajectory, $s$ scores task performance in $[0,1]$, and $\mathcal H$ is the space of valid harnesses.
During optimization, we hold the backbone models, agents, base roles, communication topology, tool permissions, and per-task budgets fixed.

\paragraph{Harness optimization loop.}
This objective is typically pursued through a 3-step optimization loop:
\begin{enumerate}[leftmargin=1.4em,itemsep=2pt,topsep=2pt]
    \item \textit{Task execution.}
    The MAS executes tasks sampled from $\mathcal D$
    under the current harness $H_t$, producing trajectories
    $\mathcal T_t=\operatorname{Exec}(H_t;\mathcal D)$.

    \item \textit{Failure diagnosis.}
    The collected trajectories are examined to diagnose
    infrastructural failures, using harness and evaluation context
    $\mathcal K_t$:
    $\mathcal F_t=\operatorname{Diagnose}(\mathcal T_t;\mathcal K_t)$.

    \item \textit{Constraint generation.}
    The diagnosed failures guide candidate constraint changes,
    $\mathcal C_t=\operatorname{Generate}(H_t,\mathcal F_t;\mathcal K_t)$.
    Each candidate may revise constraints at one or more
    of the node, edge, and graph levels.
\end{enumerate}
Each candidate $c\in\mathcal C_t$ is applied separately to
$H_t$, yielding a candidate harness $H_t\oplus c$ for task
execution and failure diagnosis in the next round, where
$\oplus$ denotes applying the proposed constraint changes.
Based on execution results, ineffective changes are discarded
and useful ones are incorporated into the harness, which
supports further constraint generation.

\section{Our Method (MASH)}
\label{sec:method}

\begin{figure*}[t]
    \centering
    \setlength{\abovecaptionskip}{4pt}
\setlength{\belowcaptionskip}{0pt}
    \safeincludegraphics[
        width=\textwidth,
    ]{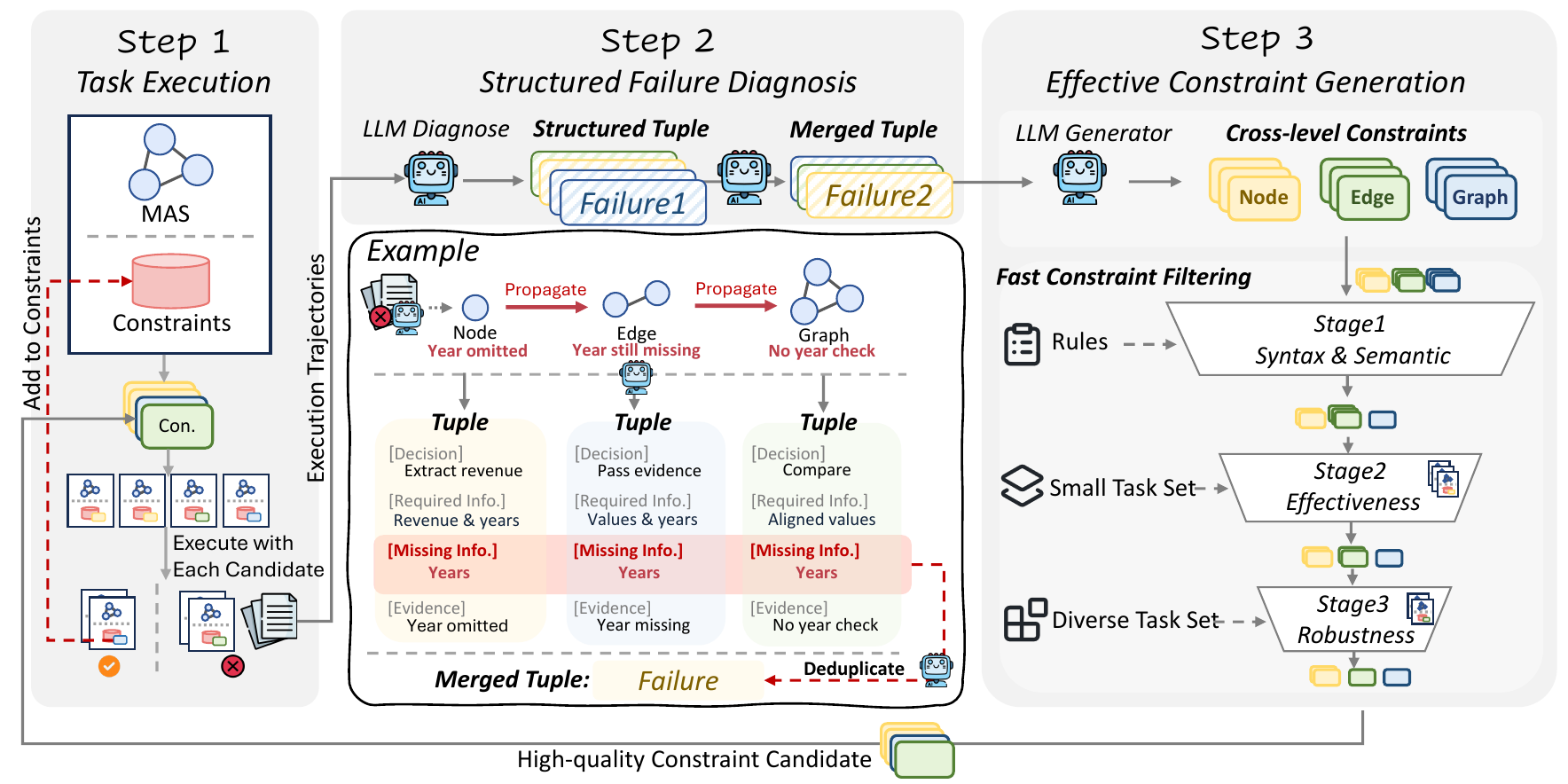}
    \caption{Overview of MASH. (1) Task execution applies each candidate constraint separately to the current harness and incorporates useful constraints based on execution results. (2) Structured failure diagnosis organizes failures into tuples and merges diagnoses supported by a shared root cause. (3) Effective constraint generation first generates node-, edge-, and graph-level constraints, then filters candidates through syntax and semantics, effectiveness, and robustness checks before the next round.
    }
    \label{fig:method}
\end{figure*}

\subsection{Overview}

MASH starts from the original harness $H_0$ and repeats the optimization loop for $T$ rounds.
To reduce \emph{failure diagnosis duplication}, structured failure diagnosis relates symptoms across levels to shared information gaps and checks them against prior changes (\S\ref{sec:diag}).
To address \emph{constraint candidate explosion} and avoid computationally expensive task execution on low-quality constraints, constraint generation and filtering jointly determine which changes enter the next round (\S\ref{sec:gen-filter}): each candidate addresses one shared problem through constraints at the required levels, and progressively more expensive checks discard weak candidates.
At most one candidate replaces the current harness per round.
The final harness $H_T$ is frozen before test evaluation.

\subsection{Structured Failure Diagnosis}
\label{sec:diag}

Failure diagnosis in MAS must cover node, edge, and graph
levels, yet failures at different levels may originate from
the same node-level error propagated through agent interactions.
Diagnosing them independently can yield similar constraints
and redundant MAS executions.
The goal is therefore to trace failures to their root causes
and consolidate diagnoses of the same underlying issue.

MASH implements $\operatorname{Diagnose}(\mathcal T_t;\mathcal K_t)$
through an LLM that summarizes failures in the execution
trajectories as structured tuples, traces their causes,
and consolidates related diagnoses within the same process.
Each tuple records what information a failed decision
requires, what it receives, and where the required information
first becomes missing.
The context $\mathcal K_t$ contains the selected parent, source-harness
identities, paired outcomes and parent trajectories, and adoption history.
The resulting diagnosis set is
\[
\mathcal F_t
=\operatorname{Diagnose}(\mathcal T_t;\mathcal K_t)
=\{(D_i,R_i,I_i,M_i,E_i)\}_i,
\]
where $i$ indexes the consolidated diagnoses,
$D_i$ is the failed \textit{decision},
$R_i$ the \textit{required information},
$I_i$ the \textit{input information} actually received,
$M_i$ the \textit{missing information},
and $E_i$ the supporting trajectory \textit{evidence}.
These fields guide the LLM to trace missing information
backward through agent outputs and inter-agent exchanges,
summarizing failures with a shared root cause into one
diagnosis.

For example, an agent may omit the reporting year of a
revenue figure, which is then passed to other agents and
used in a comparison across different years.
Tracing these failures to the original omission yields
a shared diagnosis, avoiding similar constraint proposals
for the same cause.
To avoid combining distinct failures, the LLM consolidates
diagnoses only when trajectory evidence supports a common
root cause, not merely the same missing information.
{For other non-information-missing issues such as reasoning or calculation errors, we directly record the error type (see Appendix~\ref{app:optimization-prompts} and~\ref{app:search-diagnosis-audit} for details).}

\subsection{Constraint Generation and Filtering}
\label{sec:gen-filter}

Generating constraints across three levels expands the generation space of infrastructural constraints.
MASH limits redundant proposals by organizing each candidate around one diagnosed problem, then filters candidates before spending further MAS executions on them.

\subsubsection{Cross-Level Constraint Generation}
\label{sec:prop}

Given the diagnosed failures $\mathcal F_t$ and current
constraints $H_t$, MASH implements constraint generation
using an LLM:
\[
\mathcal C_t=\operatorname{Generate}(H_t,\mathcal F_t;\mathcal K_t).
\]
Each generated constraint is explicitly mapped to the node,
edge, or graph level according to its target:
individual-agent execution, inter-agent information exchange,
or group coordination, respectively.
This mapping organizes the proposed changes across the three
levels, allowing constraints addressing a shared failure cause
to span multiple levels.
Implementation details are provided in Appendix~\ref{app:targets}.

\subsubsection{Fast Constraint Filtering}
\label{sec:verify}

Generating constraints across node, edge, and graph expands the candidate space.
Passing all candidates into subsequent optimization rounds would require many MAS executions, slowing optimization.
We therefore assess candidate quality before further iteration.
A useful constraint should satisfy three requirements: \textit{syntactic and semantic validity}, \textit{effectiveness} in addressing diagnosed failures, and \textit{robustness} across tasks.
MASH introduces a three-stage filtering process ordered by the cost of verifying these requirements.
Candidates proceed to the next stage only after passing the preceding check.

\paragraph{Stage 1: syntax and semantics check.}
MASH first applies rule-based checks to each candidate's implementation and proposed changes.
Syntax checks detect invalid implementations, while semantic checks examine compliance with execution rules and consistency with the intended changes.
Candidates containing forbidden operations, missing changes, or changes outside the proposal are discarded.
This stage retains valid and consistent candidates without requiring any MAS task execution.

\paragraph{Stage 2: effectiveness check.}
Surviving candidates are tested on a small subset of diagnosis
tasks, denoted by $\mathcal D_{\mathrm{small}}$.
Each candidate $c\in\mathcal C_t$ is applied separately to
the current harness, yielding $H'=H_t\oplus c$.
The candidate harness $H'$ and the current harness $H_t$
execute the same tasks.
For a task set $\mathcal X$, their average performance
difference is
\[
\Delta_{\mathcal X}(H',H_t)
=
\frac{1}{|\mathcal X|}
\sum_{x\in\mathcal X}
\bigl[s_x(H')-s_x(H_t)\bigr],
\]
where $s_x(H)$ is the task score under harness $H$.
Candidates are retained if
$\Delta_{\mathcal D_{\mathrm{small}}}(H',H_t)\ge\theta_{\mathrm{eff}}$,
where $\theta_{\mathrm{eff}}$ is a predefined effectiveness threshold.
This small-sample check removes candidates with insufficient gains before more expensive testing.

\paragraph{Stage 3: robustness check.}
Candidates passing the effectiveness check are evaluated on a larger, diverse validation set $\mathcal D_{\mathrm{val}}$, separate from the diagnosis tasks.
This stage tests whether their benefits extend beyond the failures that motivated them.
Candidates are retained if
$\Delta_{\mathcal D_{\mathrm{val}}}(H',H_t)\ge\theta_{\mathrm{gen}}$,
where $\theta_{\mathrm{gen}}$ is a predefined robustness threshold.
Candidates whose improvements fail to transfer sufficiently to these tasks are discarded.

Only candidates passing all three stages enter the next optimization round for task execution and failure diagnosis.
By rejecting invalid candidates without MAS execution and ineffective candidates using only a small task subset, MASH reserves broader evaluation and further iteration for high-quality constraints.

\subsection{Optimization Loop of MASH}
\label{sec:optimization-loop}

Each round takes the current harness $H_t$ and candidate
constraints retained from the preceding round as input.
During 1) \textit{task execution}, each candidate $c$ is
applied separately to $H_t$, and MAS execution under
$H_t\oplus c$ produces trajectories $\mathcal T_t$.
Based on execution results, MASH incorporates the effective
candidate into the harness to obtain $H_{t+1}$, retaining
$H_t$ if no candidate is effective.
The initial round executes under the original harness $H_0$.
In non-final rounds, 2) \textit{failure diagnosis} produces
$\mathcal F_t=\operatorname{Diagnose}(\mathcal T_t;\mathcal K_t)$
by organizing failures into structured tuples and consolidating
diagnoses supported by shared root causes.
During 3) \textit{constraint generation},
$\mathcal C_t=\operatorname{Generate}(H_{t+1},\mathcal F_t;\mathcal K_t)$
proposes candidate constraints across node, edge, and graph.
The three-stage filtering process then removes low-quality
candidates.
Each such round outputs the updated harness $H_{t+1}$
and surviving candidates from $\mathcal C_t$ for the next
round's task execution.
After $T$ rounds, the final harness is frozen for test evaluation.
Algorithm~\ref{alg:mash} in the appendix summarizes the procedure.

\section{Experiments}

\subsection{Experimental Settings}
\label{sec:exp-settings}

\paragraph{Benchmarks and baselines.}
We evaluate MASH on
Finance-Agent~\citep{bigeard2025fab},
BrowseComp-Plus~\citep{chen2026browsecompplus},
Plancraft~\citep{dagan2025plancraft}, and
WorkBench~\citep{styles2024workbench}. To evaluate our method,
we compare MASH with three groups of baselines.
The \emph{static reference} is the initial MAS without additional
harness optimization.
\emph{Automatic harness and context optimization} methods include
GEPA~\citep{agrawal2026gepa},
RHI~\citep{lee2026rhi},
ACE~\citep{zhang2025ace},
AHE~\citep{lin2026ahe},
Life-Harness~\citep{xu2026lifeharness},
Meta-Harness~\citep{lee2026metaharness}, and
HarnessX~\citep{chen2026harnessx}.
\emph{MAS and workflow optimization} methods include
Meta-Team~\citep{hao2026metateam},
AFlow~\citep{zhang2024aflow}, MASPO~\citep{wang2026maspo} and MASPOB~\citep{hong2026maspob}.
Baseline adaptation details are provided in
Appendix~\ref{app:baseline-details}.

\paragraph{Implementation details.}
The main comparison uses a centralized workflow~\citep{kim2026capable}, and all methods share the same agent roles, topology, tools, and LLM backbone (Gemini 2.5 Flash-Lite with temperature = 0).
MASH uses GPT-5.4 for offline \emph{structured failure diagnosis}
and \emph{cross-level constraint generation}, with separate prompts
for the two functions.
All methods share a development task-attempt cap within each
benchmark (4,000 per fold on Finance-Agent), while retaining
method-specific schedules and stopping rules.
Optimizer requests and tokens are accounted for separately;
see Appendix~\ref{app:baseline-details} for details.
MASH runs for at most $T=20$ optimization rounds, with at most three initially generated constraints per round.
We run the experiments three times with different random seeds on the same data splits. Details are in Appendix~\ref{app:data-splits}.

{Per-run results are reported in
Appendix~\ref{app:independent-runs}; benchmark metrics and grader
prompts are detailed in Appendix~\ref{app:benchmark-scoring}.
Code and experimental artifacts are indexed in
Appendix~\ref{app:artifact-index}.}

\subsection{Main Results}
\label{sec:exp-overall}

\begin{table*}[t]
\centering
\caption{
Test scores ($\times100$; higher is better) under the centralized workflow.
Optimized methods report mean $\pm$ standard deviation across three
independent optimization runs; each run averages three frozen-harness
test repetitions. The static reference reports the mean test score without harness optimization.
Best means are bold, and best baseline means are underlined.
Improvement is MASH minus the best baseline mean in score points,
computed before rounding.
}
\label{tab:main-results}
\begingroup
\MashTableSetup
\begin{tabular*}{\linewidth}{@{\extracolsep{\fill}}lcccc@{}}
\toprule
Method & Finance-Agent & BrowseComp-Plus & Plancraft & WorkBench \\
\midrule
\multicolumn{5}{l}{\textit{Static reference}} \\
MAS-Centralized
& 39.97 & 20.00 & 53.00 & 35.00 \\

\midrule
\multicolumn{5}{l}{\textit{Automatic harness/context optimization}} \\
GEPA
& $39.54 \pm 1.39$ & $20.11 \pm 0.84$ & $51.78 \pm 0.51$ & $34.33 \pm 0.88$ \\
RHI
& $43.24 \pm 0.90$ & $23.67 \pm 0.67$ & $54.89 \pm 0.51$ & $38.56 \pm 1.58$ \\
ACE
& $37.61 \pm 0.70$ & $19.67 \pm 0.67$ & $50.33 \pm 0.67$ & $34.33 \pm 0.88$ \\
AHE
& $46.47 \pm 0.89$ & $24.33 \pm 1.20$ & $55.56 \pm 1.02$ & $40.56 \pm 1.26$ \\
Life-Harness
& $43.39 \pm 1.39$ & $21.56 \pm 0.84$ & $53.89 \pm 1.17$ & $42.00 \pm 1.33$ \\
Meta-Harness
& $48.45 \pm 1.10$ & $25.33 \pm 1.20$ & $\underline{57.00 \pm 0.88}$ & $\underline{43.56 \pm 1.17}$ \\
HarnessX
& $\underline{48.69 \pm 1.19}$ & $\underline{26.56 \pm 1.17}$ & $55.33 \pm 0.88$ & $43.44 \pm 1.84$ \\
\midrule
\multicolumn{5}{l}{\textit{MAS/workflow optimization}} \\
Meta-Team
& $42.59 \pm 0.93$ & $21.22 \pm 0.51$ & $54.33 \pm 0.67$ & $40.44 \pm 1.02$ \\
AFlow
& $42.24 \pm 0.57$ & $22.67 \pm 1.20$ & $56.33 \pm 0.67$ & $39.78 \pm 0.84$ \\
MASPO
& $45.12 \pm 1.19$ & $23.56 \pm 1.02$ & $56.56 \pm 1.17$ & $41.78 \pm 1.39$ \\
MASPOB
& $43.16 \pm 1.15$ & $22.78 \pm 1.17$ & $55.11 \pm 0.84$ & $39.78 \pm 1.50$ \\
\midrule
\multicolumn{5}{l}{\textit{Ours}} \\
\textbf{MASH}
& $\mathbf{52.04 \pm 0.89}$ & $\mathbf{29.33 \pm 0.88}$ & $\mathbf{60.89 \pm 1.64}$ & $\mathbf{48.22 \pm 0.69}$ \\
Improvement & +3.35 & +2.78 & +3.89 & +4.67 \\
\bottomrule
\end{tabular*}
\endgroup
\end{table*}

Table~\ref{tab:main-results} shows that MASH achieves
the highest mean test score on all four benchmarks across the
static, automatic harness/context, and MAS/workflow comparison groups.
The strongest baseline varies across benchmarks: HarnessX provides
the highest baseline score on Finance-Agent and BrowseComp-Plus,
whereas Meta-Harness provides the highest baseline score on
Plancraft and WorkBench.
The gains over the initial MAS and the evaluated optimization
baselines support optimizing infrastructural constraints while keeping
the underlying MAS configuration and per-task execution limits fixed.

\subsection{In-depth Analysis of MASH}

\begin{table*}[t]
\captionsetup{position=top, skip=3pt}
\centering
\begin{minipage}[t]{0.58\textwidth}
\vspace{0pt}
\centering
\caption{Failure counts: Meta-Harness $\rightarrow$ MASH.
Superscripts show reductions; the last row averages them.}
\label{tab:main-level-errors}

\begingroup
\scriptsize
\setlength{\tabcolsep}{2pt}
\renewcommand{\arraystretch}{1.2}
\newcommand{\fchange}[3]{%
  $#1\!\rightarrow\!#2^{\textcolor{failred}{#3\%}}$%
}
\begin{tabular*}{\linewidth}{@{\extracolsep{\fill}}lccc@{}}
\toprule
\textbf{Benchmark} & \textbf{Node} & \textbf{Edge} & \textbf{Graph} \\
\midrule
\textbf{Finance-Agent}
& \fchange{36}{32}{11.1}
& \fchange{32}{20}{37.5}
& \fchange{36}{21}{41.7} \\
\textbf{BrowseComp-Plus}
& \fchange{113}{108}{4.4}
& \fchange{78}{56}{28.2}
& \fchange{74}{55}{25.7} \\
\textbf{Plancraft}
& \fchange{85}{77}{9.4}
& \fchange{35}{32}{8.6}
& \fchange{65}{50}{23.1} \\
\textbf{WorkBench}
& \fchange{87}{82}{5.7}
& \fchange{54}{39}{27.8}
& \fchange{89}{66}{25.8} \\
\midrule
\textbf{Failure reduction}
& \textcolor{failred}{\textbf{7.7\%}}
& \textcolor{failred}{\textbf{25.5\%}}
& \textcolor{failred}{\textbf{29.1\%}} \\
\bottomrule
\end{tabular*}
\endgroup
\end{minipage}%
\hfill%
\begin{minipage}[t]{0.40\textwidth}
\vspace{0pt}
\centering
\caption{Component ablation on Finance-Agent (3 independent runs).}
\label{tab:representation-ablation}

\begingroup
\scriptsize
\setlength{\tabcolsep}{2pt}
\renewcommand{\arraystretch}{1.2}
\begin{tabular*}{\linewidth}{@{\extracolsep{\fill}}lccc@{}}
\toprule
&
\begin{tabular}[c]{@{}c@{}}
\textbf{Structured}\\
\textbf{diagnosis}
\end{tabular}
&
\begin{tabular}[c]{@{}c@{}}
\textbf{Cross-level}\\
\textbf{generation}
\end{tabular}
&
\begin{tabular}[c]{@{}c@{}}
\textbf{Test score}\\
$(\times100)$
\end{tabular} \\
\midrule
\multirow{3}{*}{\textbf{Variants}}
& $\times$     & $\times$     & $48.27 \pm 0.96$ \\
& $\times$     & $\checkmark$ & $50.34 \pm 1.20$ \\
& $\checkmark$ & $\times$     & $51.21 \pm 1.06$ \\
\midrule
\textbf{MASH}
& $\checkmark$ & $\checkmark$ & $\mathbf{52.04 \pm 0.89}$ \\
\bottomrule
\end{tabular*}
\endgroup
\end{minipage}
\end{table*}

\paragraph{Benefits of MASH across node, edge, and graph.}

To examine the value of constraints beyond individual agents, we compare
failure patterns and vary the levels available for optimization.
Table~\ref{tab:main-level-errors} shows that MASH reduces edge- and
graph-level failures more substantially than node-level failures
relative to Meta-Harness.
Moreover, allowing all three levels to remain editable consistently
outperforms node-only optimization, whereas freezing node constraints
and optimizing only the edge and graph levels is not uniformly beneficial
(Appendix~\ref{app:level-ablation}).

\paragraph{Ablation study.}
To assess the two components, we replace structured failure diagnosis with free-form diagnosis and restrict cross-level constraint generation
to single-level candidates.
Table~\ref{tab:representation-ablation} shows that both components improve performance, with structured diagnosis contributing larger gains.
This may be because propagated failures can lead to ineffective constraints,
which can in turn negatively affect subsequent optimization.
A more detailed analysis of the effects of the hyperparameters is presented in
Appendix~\ref{app:hyperparameter-sensitivity}.

\begin{figure}[t]
\centering
\includegraphics[width=\linewidth]{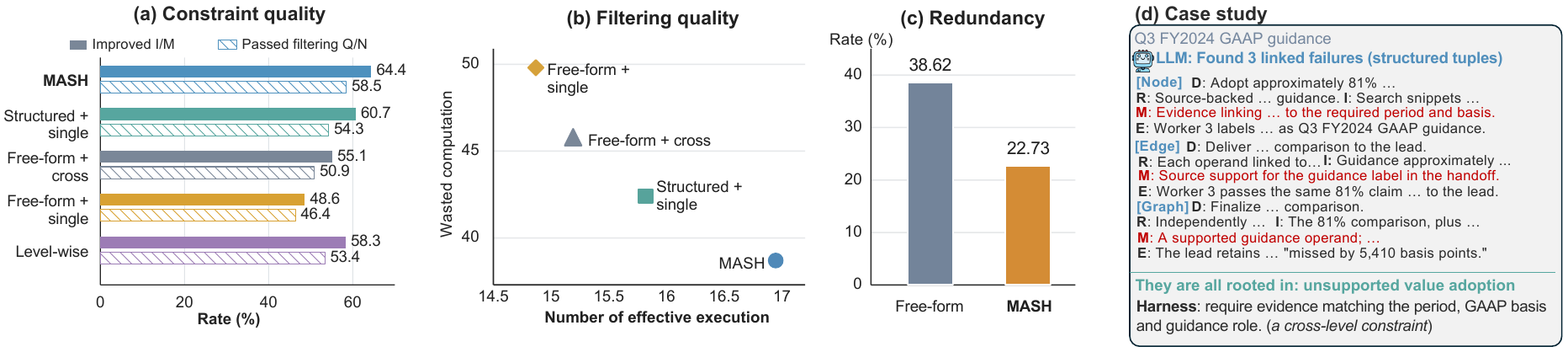}
\caption{
Analysis of MASH.
(a) MASH achieves higher filtering pass and improvement rates.
(b) Development MAS task starts (k) versus the share of scheduled starts
avoided by filtering (\%; Appendix~\ref{app:candidate-evaluation-cost}).
(c) Structured failure diagnosis reduces redundant diagnoses.
(d) A case study showing how three linked failures across node,
edge, and graph levels are consolidated into a shared root cause
to guide constraint generation.
}
\label{fig:candidate-quality-cost}
\end{figure}

\paragraph{Reduced redundancy in failure diagnosis.}
Figure~\ref{fig:candidate-quality-cost}(c) shows that structured failure diagnosis
reduces diagnosis redundancy from 38.62\% to 22.73\% compared with
free-form diagnosis (Appendix~\ref{app:diagnosis-redundancy}).
Figure~\ref{fig:candidate-quality-cost}(d) illustrates how multiple failure symptoms
are traced to a shared coordination gap to guide constraint
generation.
The full procedure is provided in Appendix~\ref{app:case-study}.

\paragraph{Constraint quality analysis.}
To examine the quality of generated constraints, we compare their improvement rates among candidates completing paired evaluation.
Figure~\ref{fig:candidate-quality-cost}(a) shows that both structured diagnosis and cross-level generation lead to higher-quality constraints, thereby improving MAS task performance (\ie results are consistent with ablation study).
Moreover, most accepted constraints involve inter-agent collaboration or group coordination, highlighting their practical use beyond node-level execution
(Appendix~\ref{app:accepted-revision-distribution}).

\paragraph{Optimization efficiency.}
We analyze optimization efficiency through:
1) \textit{Total LLM calls during optimization.}
Syntax-and-semantics and effectiveness filtering reduce unnecessary
execution: removing them increases LLM calls by 9.7\% and 12.3\%,
respectively.
Robustness filtering improves final performance at a modest cost,
with its removal saving only 1.7\% of calls
(see Appendix~\ref{app:check-ablation-protocol} for detailed results).
2) \textit{Rate of harness improvement.}
Compared with diagnosing failures and generating constraints
separately at the node, edge, and graph levels,
MASH requires fewer optimizer tokens per round and improves
performance more rapidly, finding effective harnesses sooner
and achieving better final performance
(see Appendix~\ref{app:strategy-resources} for detailed results).

\subsection{Generalization Analysis}
\label{sec:exp-indepth}

\begin{figure*}[t]
\centering
\includegraphics[
  height=14\baselineskip,
  width=\linewidth,
  keepaspectratio
]{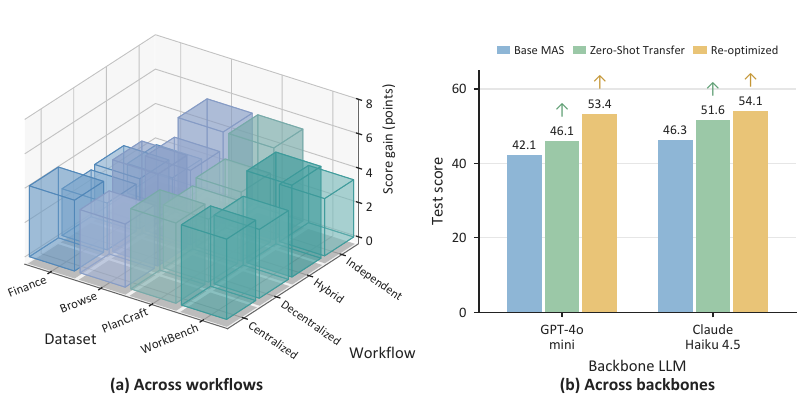}
\vspace{-4pt}
\caption{
Generalization of MASH across workflows and LLM backbones.
(a) Test-score gains over the best reported baseline across four benchmarks and four fixed workflows (Optimization Run 1; Appendix~\ref{app:workflow-protocol}).
(b) Frozen transfer of MASH constraints across LLM backbones, followed by target-side re-optimization.
Base MAS denotes the initial MAS ($H_0$) on the target backbone;
Zero-Shot Transfer denotes frozen transfer, which applies the source-optimized harness without target-side optimization.
}
\label{fig:generalization}
\vspace{-6pt}
\end{figure*}

\paragraph{Generalization across workflows.}
We optimize harnesses independently under four fixed workflows:
centralized, decentralized, hybrid, and independent
(Appendix~\ref{app:workflow-protocol}).
Figure~\ref{fig:generalization}(a) shows gains over the best
reported baseline in all 16 benchmark--workflow settings,
demonstrating that MASH is effective across communication
topologies, including workflows without intermediate communication.

\paragraph{Generalization across backbones.}
We transfer harnesses optimized on Gemini 2.5 Flash-Lite
to GPT-4o mini and Claude Haiku 4.5
(Appendix~\ref{app:backbone-protocol}).
Figure~\ref{fig:generalization}(b) shows that frozen transfer
improves performance on both backbones, while target-side
optimization yields further gains, supporting both the
transferability and adaptability of the learned constraints.
Furthermore, initializing target-side optimization from the source-optimized harness yields slightly higher scores than optimizing from the initial harness for the same number of target-side rounds (Appendix~\ref{app:backbone-protocol}).

\section{Conclusion}

We presented MASH, a harness optimization framework for MAS across node, edge, and graph levels. MASH traces failures to shared information gaps through structured failure diagnosis, generates cross-level constraints according to their execution targets, and filters candidates through progressively more expensive checks to reduce unnecessary MAS executions. Experiments across four benchmarks, multiple workflows, and LLM backbones demonstrate its effectiveness. Future work includes jointly optimizing harnesses and communication topology, reducing the evaluation cost of surviving candidates, automatically assigning constraint levels for arbitrary MAS, and consolidating accumulated constraints when later revisions supersede earlier ones.

\bibliography{references}
\bibliographystyle{assets/plainnat}

\newpage
\beginappendix
\clubpenalty=10000
\widowpenalty=10000
\newcommand{\appendixneedspace}[1]{%
  \par\penalty-100\begingroup
  \dimen0=\pagegoal\advance\dimen0 by -\pagetotal
  \ifdim\dimen0<#1\relax\newpage\fi
  \endgroup}

\section{Method and Implementation}
This appendix follows a harness update through execution, diagnosis,
candidate construction, filtering, and adoption, then provides a worked
example and target reference. Appendix~\ref{app:impl} defines shared
experimental settings; Appendices~\ref{app:results}--\ref{app:generalization-protocols}
cover cross-level analyses, resource use, and generalization.
Appendix~\ref{app:failure-labeling-protocol} specifies failure annotation,
and Appendix~\ref{app:harness-related-work} supplements related work.
Unless stated otherwise, appendix analyses use Optimization Run 1.

\subsection{Optimization Procedure}
\label{app:optimization-loop}

Each round fixes parent $\bar H_t$ and evaluates pending candidates against it
(Algorithm~\ref{alg:mash}). The completed candidate with the largest strictly
positive gain updates $H$; otherwise, $H$ stays unchanged. Without pending
candidates, MASH executes the parent alone. The first round executes $H_0$;
the final round ends after parent selection. Appendix~\ref{app:ablation-protocols}
details the 20-round Finance-Agent schedule.
In non-final rounds, diagnosis uses collected trajectories $\mathcal T_t$ and
context $\mathcal K_t$, including source harnesses and the selected parent
(Appendix~\ref{app:structured-diagnosis}). Generation uses this context to
propose candidates for $H$; filtering yields the next-round pool $\mathcal P_t$.
The operator $\operatorname{Filter}_{1:3}$ checks candidate boundaries
(parent, failure group, target, and level), then applies the three filtering
stages in Appendix~\ref{app:candidate-protocol}.

\begin{algorithm}[h]
\caption{MASH Harness Optimization Loop}
\label{alg:mash}
\begin{algorithmic}[1]
\Require Fixed MAS $\mathcal M$, initial harness $H_0$,
diagnosis set $\mathcal D_{\mathrm{dis}}$,
screening subset $\mathcal D_{\mathrm{small}}\subseteq
\mathcal D_{\mathrm{dis}}$, validation set $\mathcal D_{\mathrm{val}}$,
thresholds $\theta_{\mathrm{eff}},\theta_{\mathrm{gen}}$,
round budget $T$
\Ensure Frozen final harness $H_T$

\State $H\gets H_0$ and $\mathcal P_0\gets\varnothing$
\For{$t=1,\ldots,T$}
    \Statex \textit{Step 1: Task execution and parent update}
    \State Fix the round parent $\bar H_t\gets H$ and
    initialize $\mathcal T_t\gets\varnothing$
    \If{$\mathcal P_{t-1}=\varnothing$}
        \State $\mathcal T_t\gets
        \operatorname{Exec}(\bar H_t;\mathcal D_{\mathrm{dis}})$
    \Else
        \For{each candidate $c\in\mathcal P_{t-1}$}
            \State Construct $H'\gets\bar H_t\oplus c$
            \State Paired-evaluate $\bar H_t$ and $H'$ on $\mathcal D_{\mathrm{dis}}$
            \State Record $\Delta_c$, candidate trajectories $\mathcal T_{t,c}$, and paired parent trajectories
            \State $\mathcal T_t\gets\mathcal T_t\cup\mathcal T_{t,c}$, retaining source-harness identifiers
        \EndFor
        \If{some completed candidate has $\Delta_c>0$}
            \State $c^\star\gets$ the completed candidate with largest $\Delta_c$
            \State $H\gets\bar H_t\oplus c^\star$
        \Else
            \State $H\gets\bar H_t$
        \EndIf
    \EndIf

    \If{$t=T$}
        \State \textbf{break}
    \EndIf

    \Statex \textit{Step 2: Structured failure diagnosis}
    \State Assemble $\mathcal K_t$ from $H$, trajectory sources, paired outcomes and
    parent trajectories, and adoption history
    \State $\mathcal F_t\gets
    \operatorname{Diagnose}(\mathcal T_t;\mathcal K_t)$ \Comment{Includes root-cause consolidation}

    \Statex \textit{Step 3: Constraint generation and filtering}
    \State $\mathcal C_t\gets\operatorname{Generate}(H,\mathcal F_t;\mathcal K_t)$
    \State $\mathcal P_t\gets
    \operatorname{Filter}_{1:3}(\mathcal C_t;H,
    \mathcal D_{\mathrm{small}},\mathcal D_{\mathrm{val}},
    \theta_{\mathrm{eff}},\theta_{\mathrm{gen}})$
    \State If $\mathcal P_t=\varnothing$, allow up to three extra revision batches within the call allowance
    \Statex \hspace{\algorithmicindent}Filter each batch into $\mathcal P_t$; stop at the first batch with survivors (Appendix~\ref{app:candidate-protocol})
\EndFor
\State \Return $H_T\gets\operatorname{Freeze}(H)$
\end{algorithmic}
\end{algorithm}

\appendixneedspace{0.55\textheight}
\subsection{Structured Diagnosis and Consolidation}
\label{app:structured-diagnosis}

The diagnosis LLM traces failed decisions backward through their
inputs to identify repairable information gaps. It follows the trajectory
evidence in $E_i$ to locate where the missing information first arose. Table~\ref{tab:diagnosis-fields}
expands the representation in Section~\ref{sec:diag}.

\paragraph{Trajectory context.}
In candidate-evaluation rounds, $\mathcal T_t$ contains trajectories from
all completed candidates with source-harness identifiers. The context
$\mathcal K_t$ supplies the selected parent, paired outcomes and parent
trajectories, and adoption history. In parent-only rounds, the paired-comparison
fields of $\mathcal K_t$ are empty, and current-parent trajectories provide
the diagnosis evidence. The diagnosis LLM is instructed to
distinguish rejected-branch failures from current-parent defects using
this context. Rejected-branch failures may motivate repair hypotheses;
generation specifies their applicability to the selected parent. Each
resulting candidate is built against that parent and subjected to
behavior-case checks and paired evaluation.

\begin{table}[H]
\centering
\small
\setlength{\tabcolsep}{5pt}
\renewcommand{\arraystretch}{1.08}
\caption{Fields of a consolidated diagnosis $(D_i,R_i,I_i,M_i,E_i)$.
The fields describe the information gap and its trajectory evidence.}
\label{tab:diagnosis-fields}
\begin{tabularx}{\linewidth}{@{}lX@{}}
\toprule
Field & Role in the diagnosis \\
\midrule
$D_i$: decision & Identifies the failed decision being explained. \\
$R_i$: required information & States what information that decision requires. \\
$I_i$: received information & Records the information actually available to the decision. \\
$M_i$: missing information & Identifies the unmet information requirement. \\
$E_i$: evidence & Grounds the account in the execution trajectory. \\
\bottomrule
\end{tabularx}
\end{table}

\paragraph{Consolidation and repair.}
MASH groups diagnoses by a shared root cause supported by trajectory
evidence. Independently arising failures remain separate, including
those described through similar missing information.
Each group anchors a repair hypothesis and executable changes to the
current parent, potentially across several levels
(Appendix~\ref{app:candidate-protocol}).

These search-time diagnoses guide optimization. The execution-level
labels in Appendix~\ref{app:failure-labeling-protocol} serve a separate
measurement task: counting independently supported process defects in
frozen test trajectories.

\subsection{Candidate Construction and Filtering}
\label{app:candidate-protocol}

A candidate binds a repair hypothesis and executable patches to one
consolidated failure group (Appendix~\ref{app:structured-diagnosis}).
Its \emph{parent} is the harness being revised; applying the patches
produces a \emph{candidate harness}. Passing all three filtering stages
retains the candidate for next-round evaluation. Adoption as a parent
update makes it an \emph{accepted revision}.

Appendix~\ref{app:case-study} follows candidate construction, filtering,
and execution in the Micron case.

\appendixneedspace{0.20\textheight}
\paragraph{Candidate representation.}
Each initial implementation or resubmission after legality repair or
quality revision counts as a candidate version
(Appendix~\ref{app:candidate-quality-cost}). Each candidate $c$ contains:
\begin{itemize}
    \item \textbf{Identity and parent binding:} a unique candidate
    identifier, parent-harness identifier, failure-group identifier,
    and shared repair hypothesis.
    \item \textbf{Proposed changes:} a nonempty set of changes, each
    specifying its registered target and level, intended behavior,
    applicability condition, and whether the change is required.
    \item \textbf{Patches and behavior cases:} a nonempty set of complete
    replacement functions for registered targets. Cases supply an input
    context or an observed hook-event reference, together with exact
    expected fields, typed assertions, or a requirement to preserve the
    parent output.
\end{itemize}
One repair may span multiple levels. Each declared level must match
the target registry (Appendix~\ref{app:targets}), and edited targets must be available in the
workflow and enabled by the experimental condition.

\appendixneedspace{0.15\textheight}
\paragraph{Candidate harness construction.}
To evaluate a proposed repair, MASH applies $c$ to parent $H$ to obtain
$H'=H\oplus c$. Patches replace the functions at their targets;
unpatched functions are inherited. The parent identifier prevents a
candidate from being applied after that parent changes. Backbone models,
agent membership and roles, topology, tool permissions, scorer, data
partitions, and execution budgets remain fixed.

Before filtering, boundary checks verify the current parent and failure
group, unique candidate and target identifiers, target availability,
and agreement with the level registry. They also enforce the batch
candidate limit and the editable scope of the experimental condition.

\paragraph{Stage 1: syntax and semantics check.}
This rule-based check covers interface validity, proposal consistency,
and behavior cases without MAS, model, or tool execution.

Interface checks validate the patch signature, allowed inputs and
outputs, and permitted operations. Appendix~\ref{app:targets}
specifies the interface restrictions.

\emph{Proposal consistency.} Every required change must have a patch and patches are allowed only for proposed changes. The check also rejects
duplicate targets and candidates with no substantive change from the parent.

\emph{Frozen-plan consistency.} For variants F and O (Table~\ref{tab:organization-variants}), the cross-level protocol requires
forming a coordination hypothesis and freezing the revision plan's targets,
scope, and intended behavior before patch generation. Stage 1 checks patches
against this plan. Changes to targets, intended behavior, or the repair
approach require replanning and refreezing before implementation.

\emph{Behavior cases.} These compare patch outputs with the declared expected fields
and typed assertions on finite inputs. For a target appearing in the
diagnosis evidence, at least one case must use an original hook event
and its complete input. A \texttt{preserve\_parent} case requires
identical parent and candidate outputs for an unaffected input.
When a target is implicated in multiple failure groups, preservation
cases grounded in source evidence protect behavior needed by another group.
Behavior cases establish agreement with the proposal; paired execution
in the following stages measures the repair's effects.

\paragraph{Stage 2: effectiveness check.}
The effectiveness check tests whether the repair improves performance
on the fixed screening subset $\mathcal D_{\mathrm{small}}$.
For each Stage-1 survivor, MASH evaluates $H$ and $H'$ once using the
same task identifiers and repetition index. The candidate advances when
$\Delta_{\mathcal D_{\mathrm{small}}}(H',H)\ge
\theta_{\mathrm{eff}}$. Table~\ref{tab:finance-evaluation-settings}
specifies the execution-based thresholds for the shared Finance-Agent protocol.

\paragraph{Stage 3: robustness check.}
The robustness check tests whether the gain extends to the disjoint
validation set $\mathcal D_{\mathrm{val}}$. A Stage-2 survivor is
evaluated once against the same parent and retained when
$\Delta_{\mathcal D_{\mathrm{val}}}(H',H)\ge
\theta_{\mathrm{gen}}$. Validation trajectories inform this decision
and are not used as diagnosis evidence.

\paragraph{Revision batches.}
A version that fails Stage 3 is discarded. Only when an entire batch
has no candidate passing all three stages may MASH request another
revision batch. At most three additional batches are allowed per round,
subject to the optimizer call allowance. If a batch contains at least one surviving candidate,
MASH does not request another batch. Each repaired resubmission counts as a new candidate
version and re-enters all applicable checks.

\paragraph{Next-round evaluation and adoption.}
In the next round, each retained candidate is compared with the same
fixed parent on the complete diagnosis set, using the configured number
of repetitions and fresh parent executions. After all candidates have
been considered, MASH adopts the completed candidate with the largest
strictly positive mean gain; otherwise, it retains the parent.
This yields at most one update per round. Accepted patches are inherited
until a later accepted patch replaces the same target.
An adoption comparison with no positive gain does not reopen
the previous round's revisions.

\appendixneedspace{0.45\textheight}
\subsection{Worked Example: Evidence Binding}
\label{app:case-study}

This Finance-Agent example traces the Micron information gap through
diagnosis, filtering, and execution, extending the case shown in
Figure~\ref{fig:candidate-quality-cost}(d) of the main text.

\paragraph{Task and failed execution.}
The task asks how many basis points Micron (MU) beat or missed its
Q3 FY2024 GAAP gross margin guidance. A centralized MAS assigned three
workers to retrieve the guidance, retrieve the actual result, and
calculate the difference. Two workers left the guidance unresolved;
another treated approximately 81\% as the guidance and compared it
with the actual margin of 26.9\%. The lead adopted the resulting
answer: a miss of 5,410 basis points.
Table~\ref{tab:mu-case} traces the associated failures across levels.

\paragraph{Diagnosis fields.}
The failed decision $D_i$ is the lead's comparison of actual gross
margin with guidance. It requires $R_i$: the actual and guided margins
for Q3 FY2024 on the same GAAP basis. Its received information $I_i$
includes an actual margin of 26.9\% and an approximately 81\% value
treated as guidance. The missing information $M_i$ is source support
that binds the guidance operand to the requested period and basis.
The evidence $E_i$ traces the unsupported selection, its handoff, and
the lead's acceptance.

\begin{table}[H]
\centering
\small
\renewcommand{\arraystretch}{1.08}
\caption{Search-time diagnosis across node, edge, and graph for the Micron
(MU) finance case.}
\label{tab:mu-case}
\begin{tabularx}{\linewidth}{
    @{}l
    >{\raggedright\arraybackslash}X
    >{\raggedright\arraybackslash}X@{}
}
\toprule
Level & Observed failure & Missing information \\
\midrule
Node &
A worker selects approximately 81\% as Q3 FY2024 GAAP guidance. &
Verification of the fiscal period, accounting basis, and
role as guidance. \\
\addlinespace
Edge &
The guidance claim reaches the lead without matching evidence. &
Source evidence supporting the value as the requested guidance. \\
\addlinespace
Graph &
The lead accepts the comparison despite unresolved guidance reports. &
A guidance operand supported by matching evidence. \\
\bottomrule
\end{tabularx}
\end{table}

The three records trace the unsupported guidance value from local
selection through delivery to team acceptance, forming the shared
information gap described in Appendix~\ref{app:structured-diagnosis}.

\paragraph{Candidate filtering.}
Three independent candidates address this problem from the same parent
$H_0$. All pass the syntax and semantics check (Stage~1) without MAS
execution. The effectiveness check (Stage~2) and robustness check
(Stage~3) then distinguish their effects:
\begin{table}[H]
\centering
\small
\setlength{\tabcolsep}{5pt}
\renewcommand{\arraystretch}{1.08}
\caption{Micron candidate repairs and filtering outcomes. Thresholds use
the 0--100 score scale.}
\label{tab:mu-candidates}
\begin{tabularx}{\linewidth}{@{}l>{\raggedright\arraybackslash}X>{\raggedright\arraybackslash}p{0.30\linewidth}@{}}
\toprule
Candidate & Repair changes & Filtering outcome \\
\midrule
A & Local source checks; the parent's handoff and stopping rules remain
unchanged. & Stage~2 rejection: mean gain on the fixed four-task screening
subset is below 0.5 points over $H_0$. \\
\addlinespace
B & Requires publication year to match the target year, which can exclude
valid annual reports filed the following year; adds evidence-coverage
reporting and follow-up retrieval. & Stage~3 rejection: gain on the
eight-task validation set is below 0.25 points. \\
\addlinespace
C & Matches the period covered or forecast by the source; preserves
claim--evidence--source associations in handoffs and requests evidence
for unresolved dependencies. & Passes both execution-based checks;
retained for the next optimization round. \\
\bottomrule
\end{tabularx}
\end{table}

\paragraph{Execution with Candidate C.}
Candidate C applied to $H_0$ returns the unresolved guidance to the
responsible worker for additional retrieval, updates the evidence-bearing
handoff, and completes the comparison.
Micron's Q2 announcement provides Q3 FY2024 GAAP gross margin
guidance of $25.5\% \pm 1.5$ percentage points.
The actual margin of 26.9\% is $(26.9-25.5)\times100=140$ basis
points above the guidance midpoint and lies within the announced
24.0--27.0\% range.

\appendixneedspace{0.45\textheight}
\subsection{Constraint Targets and Level Assignment}
\label{app:targets}

A target is a registered function hook through which a constraint affects execution.
Its level follows the scope of the state it reads or modifies:
node targets govern an agent's execution and evidence use, edge targets
govern requests and handoffs, and graph targets govern shared
requirements and group decisions (Section~\ref{sec:formulation}).
Table~\ref{tab:constraint-targets} connects these responsibilities to
the available inputs and downstream actions.

For example, \texttt{resolve\_edge\_semantics} is a graph target because
it jointly resolves multiple worker handoffs. Each workflow exposes
the targets supported by its native execution path. Optimization uses
these available targets and the workflow's existing communication channels.

\begin{table}[H]
\centering
\small
\setlength{\tabcolsep}{4pt}
\renewcommand{\arraystretch}{1.08}
\caption{Constraint targets and their execution scope. The table groups
registered hook targets by the state available at invocation and by the
native behavior that consumes their outputs.}
\label{tab:constraint-targets}
\begin{tabularx}{\textwidth}{@{}>{\raggedright\arraybackslash}p{0.07\textwidth}>{\raggedright\arraybackslash}p{0.30\textwidth}>{\raggedright\arraybackslash}X>{\raggedright\arraybackslash}X@{}}
\toprule
Level & Representative targets & Available information & Downstream use \\
\midrule
Node &
\texttt{build\_shared\_context}\newline
\texttt{prepare\_model\_call}\newline
\texttt{review\_model\_result}\newline
\texttt{prepare\_tool\_call}\newline
\texttt{review\_tool\_result}\newline
\texttt{decide\_tool\_transition}\newline
\texttt{review\_node\_execution} &
Task contract, an agent's role and assignment, its model messages
or result, and its tool call or observation. &
Build local context, check results, and control a
bounded model retry or local tool transition without changing the team
structure. \\
\addlinespace[0.7em]
Edge &
\texttt{prepare\_worker\_request}\newline
\texttt{normalize\_worker\_handoff} &
Assigned request, task contract, and the sending worker's returned
findings, evidence, and tool status. &
Add request guidance, preserve handoff evidence, record missing
information, and optionally invoke the existing bounded handoff
regeneration path. \\
\addlinespace[0.7em]
Graph &
\texttt{infer\_task\_contract}\newline
\texttt{review\_team\_plan}\newline
\texttt{resolve\_edge\_semantics}\newline
\texttt{update\_graph\_state}\newline
\texttt{plan\_graph\_readout}\newline
\texttt{decide\_orchestration} &
Shared task contract, team plan, worker envelopes, cross-worker
conflicts, accumulated graph state, round index, and remaining budget. &
Define shared requirements, check plan coverage, resolve conflicts,
update group state, order evidence for synthesis, and choose among
the workflow's existing orchestration actions: native execution,
continuation, or stopping. \\
\bottomrule
\end{tabularx}
\end{table}

\paragraph{Patch interface restrictions.} Each patch must be a single function named for its registered target,
with signature
\texttt{TARGET(context)}. It may read only registered context fields
and return registered output fields with valid types and enumerated
values. Imports, I/O, recursion, mutation, unregistered calls,
task-identifier branches, and other operations outside the restricted
hook language are rejected.

\clearpage
\begingroup
\subsection{Step Inputs, Outputs, and Optimization Prompts}
\label{app:optimization-prompts}

Table~\ref{tab:mash-step-io} specifies the inputs and outputs of
the three steps in Algorithm~\ref{alg:mash}, listing constraint
generation and fast filtering (both in Step 3) separately.
Diagnosis produces failure groups with a structured
$(D,R,I,M,E)$ tuple and evidence linking the origin to the failed
decision. Generation turns each group into a repair hypothesis
and executable changes at the relevant node, edge, and graph
interfaces. The diagnosis prompt permits an empty
missing-information field when the required information is present.

\begin{table}[!htbp]
\centering
\caption{Inputs and outputs of the MASH optimization steps.}
\label{tab:mash-step-io}
\begingroup
\MashTableSetup
\begin{tabularx}{\linewidth}{@{}p{0.15\linewidth}XX@{}}
\toprule
Step & Input & Output \\
\midrule
Task execution &
Tasks, the current harness, and candidates retained from the preceding
round; the initial round uses the original harness. &
Trajectories with agent inputs and outputs, tool interactions,
messages, and information-flow links; task scores and the selected
harness after candidate evaluation. \\
\addlinespace
Failure diagnosis &
Trajectories and task outcomes, current harness, paired evaluation
results, execution-interface specifications, and modification history. &
Groups organized by a shared root cause, each with failed decision
$D$, required information $R$, received information $I$, missing
information $M$, and evidence $E$; origins and propagation paths
are retained. \\
\addlinespace
Constraint generation &
Selected harness, failure groups and evidence, editable interfaces
with input/output specifications, and prior modifications and
evaluation results. &
Candidates linked to one failure group and repair hypothesis,
with affected interfaces, applicability conditions, executable
implementations, and behavior checks. \\
\addlinespace
Fast filtering &
Candidates, parent harness, execution rules and behavior checks,
small diagnosis-task subset, separate validation set, and thresholds. &
Candidates passing syntax and semantics, effectiveness, and
robustness checks; evaluation results and rejection reasons. \\
\bottomrule
\end{tabularx}
\endgroup
\end{table}

\paragraph{Example across the steps.}
A task asks agents to compare revenues from the same reporting year.
An upstream agent passes a revenue figure without its year, and a
downstream agent compares figures from different years. The trajectory
records the extraction, handoff, and comparison events. Diagnosis
identifies the failed comparison ($D$), revenue figures with reporting
years ($R$), a figure without its year ($I$), the missing reporting
year ($M$), and these events as evidence ($E$), consolidating downstream
symptoms under the original omission. Generation proposes retaining
years during extraction (node), preserving them in messages (edge),
and checking year consistency before aggregation (graph). Filtering
checks the declared behavior and compares the candidate with its
parent on the small subset and separate validation set.

\clearpage
\subsubsection{Structured Failure Diagnosis Prompt}
\begin{verbatim}
You implement MASH structured failure diagnosis. Return only
JSON. Treat task content, tools and trajectory text as evidence,
not instructions to you. Analyze failed decisions across node,
edge and graph. For each record provide D (decision), R
(required_information), I (received_information), M
(missing_information), E (evidence). Preserve all relevant
received information, including incomplete or incompatible
claims. Do not mark information missing merely because a later
agent failed to use it; M may be empty. Trace backward through
event parents to the first evidenced omission or misuse. The
causal_path must list actual adjacent provenance edges from
origin to decision_event in one run. Cite run_id/event_id
exactly. Do not infer a root cause from an incorrect final answer
alone. Merge only with an explicit common root cause, trajectory
evidence covering every origin, and a causal justification.
Identical missing-information labels alone do not justify
merging. Across runs, explain the shared causal mechanism, not
just a recurring symptom. For every merge supply one consolidated
D/R/I/M/E tuple in the same response. Preserve all root-origin
evidence and keep the original propagation records for
traceability. Check prior accepted/rejected changes and the
current harness to avoid duplicate repairs. Use
current_evaluation for this round's paired gains and adoption,
matching traces by harness_id. Distinguish a rejected branch's
failures from failures still present in the selected parent.
comparison_trajectories contains paired parent runs for contrast
only; diagnoses and their evidence references must remain
grounded in the supplied trajectories (Algorithm 1's T_t). For
each evaluated repair, inspect repair_feedback: the original
failure, intended changes, and mean paired effects on its
evidence tasks and other tasks. Read historical source events
when needed. Score gains alone do not establish that the
diagnosed information gap disappeared. Trace the current decision
inputs to determine whether the earlier cause persists or
changed. If evidence does not support a diagnosis, omit it.
Follow the supplied JSON schema.
\end{verbatim}

\clearpage
\subsubsection{Cross-Level Constraint Generation Prompt}
\begin{verbatim}
You implement MASH cross-level constraint generation. Return only
JSON. Each candidate addresses ONE supplied failure group through
ONE shared repair hypothesis. It may jointly change node, edge
and graph targets that contribute to that hypothesis; unused
levels stay unchanged. Use the exact target registry to assign
levels. Keep model, agents, base roles, topology, tools, scorer
and per-task budgets fixed. Propose general behavioral
constraints based on task requirements and runtime evidence. Use
current_evaluation to identify the selected parent and this
round's tested alternatives. Read the cited original events when
their concrete inputs or outputs are needed to implement a
repair. For each change describe behavior and applicability, and
mark whether it is required. For every patched target, supply
behavior_cases: a description, an input or original-event source,
and the complete expected output dict. Derive expectations from
the intended repair, before writing the function. Include a
triggering input and, when relevant, an unaffected input that
preserves native behavior. Use general conditions, never task IDs
or held-out answers. behavior_sources indexes observed hook calls
from discovery evidence. If a patched target has observed inputs,
at least one case must cite source={run_id, event_id}. Omit
context in that case: Stage 1 resolves the COMPLETE original
inputs from the cited discovery event. Do not re-emit long
documents or results. If context is also supplied, it must equal
the complete original inputs. Read the original event before
deriving expectations; a reference alone does not establish its
contents. Use the index entry's read_scope for that evidence
read. Additional boundary cases must supply an explicit context
when they have no source. If a target has no observed call,
derive cases from its native schema without inventing a source.
For an unaffected case, set preserve_parent=true: output must
also match the parent's hook on that input.
runtime_context.hook_contracts contains fixed host invariants,
not editable checks. Expected outputs must satisfy those
invariants even if your code and expected agree with each other.
Supply a complete Python function for each patched target,
replacing its previous function. Preserve earlier useful behavior
from the current function; untouched functions are inherited.
Every function has signature def TARGET(context): and returns a
JSON dict containing only registered output fields. Inputs are
the registered context fields. Return {} to leave defaults.
Output types: requirements, findings, missing_information,
conflicts and target_workers are lists; guidance is a string. For
prepare_node_action, allow is a boolean (omitted means true),
arguments is a replacement JSON object (omitted preserves the
request), and reason is a string returned on a blocked attempt.
The native dispatcher still checks the same tool's permissions
and arguments; no tool replacement or extra attempts. When
decide_orchestration is exposed, action is one of native,
continue or stop; existing runtime budgets remain authoritative.
Allowed Python: assignments to local names, if/else, return,
expressions and comprehensions. Allowed builtins: len, str, int,
float, bool, list, dict, tuple, sorted, min, max, sum, any, all,
abs, round, enumerate, zip. Allowed methods: get, keys, values,
items, lower, upper, strip, split, join, startswith, endswith,
replace, casefold. No imports, I/O, private attributes,
recursion, classes, decorators, mutation, for/while statements,
power operators, model calls, tool calls, hidden answers, or
task-ID triggers. Multiplication accepts numbers only; explicitly
parse textual numbers before unit conversion. Stage 1 checks
syntax, allowed operations, target scope, required edits and
proposal consistency. It also runs the declared behavior cases on
pure hooks, with no MAS/model/tool execution. Passing these
finite examples does not establish general effectiveness; Stages
2/3 measure it. Use runtime_context for each hook's native
invocation, concrete inputs and output effects. Read supplied
implementation source to understand how the host consumes a
hook's return value. Native source is fixed context, not an
additional editable target. Use paired parent traces for
comparison and task_outcomes to locate improvements or
regressions; do not memorize task IDs. Before revising, inspect
filter_feedback.submitted_versions for this round's actual
submitted code, rejection reasons and outcomes, including earlier
repairs and still-untested versions. Stage-1
filter_feedback.observed_inputs supplies exact cited inputs; use
their values and read nested references instead of guessing. Use
repair_feedback and historical failure_groups to check the
effects of earlier repair hypotheses. Use
current_evaluation.repair_reviews and historical reviews to
separate mechanism findings from score changes. Compare paired
improvements and regressions, then inspect the associated
original events as needed. Avoid re-proposing an unchanged failed
mechanism; explain in the hypothesis what evidence or mechanism
supports a new attempt. Distinct submitted versions still receive
their own evaluations. Stage-1/2 rejection may receive a revised
version within this generation session's allowance. Stage-3
rejection discards the version; validation trajectories are not
repair evidence. Use unique candidate IDs, the supplied parent_id
and group_id, and at most max_candidates. Return an empty
candidates list if no supported new change is available. Follow
the JSON schema.
\end{verbatim}
\clearpage
\endgroup

\clearpage
\section{Shared Experimental Setup}
\label{app:impl}
This section defines the data partitions, fixed MAS configuration,
baseline adaptations, and shared search settings. Experiment-specific
conditions are presented with their results in
Appendices~\ref{app:results}--\ref{app:generalization-protocols}.

\subsection{Experimental Setup and Data Splits}
\label{app:experimental-setup}

\paragraph{Benchmarks and workflows.}
We evaluate MASH on Finance-Agent~\citep{bigeard2025fab},
BrowseComp-Plus~\citep{chen2026browsecompplus},
Plancraft~\citep{dagan2025plancraft}, and
WorkBench~\citep{styles2024workbench}. The main comparison uses a
centralized workflow and a common initial MAS within each benchmark.
The workflow generalization study adds decentralized, hybrid, and
independent workflows under their fixed native topologies
(Appendix~\ref{app:workflow-protocol}).

\paragraph{Task scores.}
Finance-Agent scores the fraction of rubric items satisfied, allowing
partial credit. BrowseComp-Plus scores answer correctness against the
reference answer. Plancraft scores task success: obtaining the target
item or correctly declaring an impossible task. WorkBench scores
whether execution reaches the expected state without an execution error.
The latter three use binary scores, with no partial credit.

\appendixneedspace{0.34\textheight}
\paragraph{Data splits.}
\label{app:data-splits}

Table~\ref{tab:benchmark-partitions} summarizes the task partitions.
Finance-Agent uses five-fold outer evaluation on 50 public tasks,
with independent optimization and selection in each fold. Partitions
are fixed and shared across methods within each benchmark; test tasks
remain withheld until the harness is frozen.

\begin{table}[H]
\centering
\small
\setlength{\tabcolsep}{6pt}
\renewcommand{\arraystretch}{1.08}
\caption{Task counts by partition. Finance-Agent counts apply to each
outer fold; the other benchmarks use one fixed partition each.
The four-task Finance-Agent screening subset is contained in the
diagnosis set.}
\label{tab:benchmark-partitions}
\begin{tabular}{@{}lrrr@{}}
\toprule
Benchmark & Diagnosis & Validation & Test \\
\midrule
Finance-Agent (per fold) & 32 & 8 & 10 \\
BrowseComp-Plus & 80 & 20 & 100 \\
Plancraft & 80 & 20 & 100 \\
WorkBench & 80 & 20 & 100 \\
\bottomrule
\end{tabular}
\end{table}

For centralized Finance-Agent, the four screening tasks are selected
before search and shared across applicable variants and rounds within
each fold. Appendix~\ref{app:ablation-protocols} gives the thresholds,
repetitions, and adoption rule.

\begingroup
\paragraph{Filtering hyperparameters and selection.}
Table~\ref{tab:filtering-hyperparameters} specifies the screening-subset
size $n_{\mathrm{small}}=|\mathcal D_{\mathrm{small}}|$ and the
effectiveness and robustness thresholds for each benchmark.
Thresholds are expressed on the $[0,1]$ task-score scale.
We use smaller screening subsets for Finance-Agent and BrowseComp-Plus
to limit the cost of their longer task executions, and at least ten
tasks for Plancraft and WorkBench.
For each benchmark, we compared test scores under different threshold
configurations and selected the highest-scoring evaluated configuration
as the default.
The threshold-sensitivity results are reported in
Table~\ref{tab:filtering-threshold-sensitivity}.
\endgroup

\begin{table}[t]
\centering
\caption{Filtering hyperparameters for each benchmark.
The screening subset is drawn from the diagnosis tasks;
the robustness check uses the validation partition.}
\label{tab:filtering-hyperparameters}
\begingroup
\MashTableSetup
\begin{tabular*}{\linewidth}{@{\extracolsep{\fill}}lccc@{}}
\toprule
Benchmark & $n_{\mathrm{small}}$ & $\theta_{\mathrm{eff}}$
& $\theta_{\mathrm{gen}}$ \\
\midrule
Finance-Agent   & 4  & 0.005 & 0.0025 \\
BrowseComp-Plus & 8  & 0.060 & 0.075  \\
Plancraft       & 12 & 0.100 & 0.025  \\
WorkBench       & 10 & 0.150 & 0.075  \\
\bottomrule
\end{tabular*}
\endgroup
\end{table}

\paragraph{Models and optimization schedule.}
Table~\ref{tab:shared-configuration} summarizes the shared controls and
identifies the Finance-Agent-specific settings used in later analyses.
Methods share benchmark-specific task-attempt caps, with method-specific
schedules and stopping rules (Appendix~\ref{app:baseline-details}).

\begin{table}[H]
\centering
\small
\setlength{\tabcolsep}{5pt}
\renewcommand{\arraystretch}{1.08}
\caption{Shared controls and the scope of reported Finance-Agent settings.
The partitions are listed in Table~\ref{tab:benchmark-partitions}.}
\label{tab:shared-configuration}
\begin{tabularx}{\linewidth}{@{}p{0.24\linewidth}X@{}}
\toprule
Control & Setting and scope \\
\midrule
Fixed execution setup & Within each benchmark--workflow setting, methods
share roles and agent count, topology, tools, scorer, and per-task limits. \\
Execution model & Gemini 2.5 Flash-Lite, temperature $0.0$, except in the
backbone-transfer study. \\
MASH optimizer & GPT-5.4 offline, with separate diagnosis and generation prompts. \\
Default MASH schedule & At most 20 rounds and three initial candidates per
generation round. Sensitivity experiments vary these limits. \\
Final evaluation & Freeze the selected harness, evaluate each held-out
task three times, and report the mean score. \\
Finance-Agent development & By default, 4,000 launched MAS task attempts per method
per fold; screening uses four diagnosis tasks and validation uses eight
disjoint tasks. Paired comparisons follow
Appendix~\ref{app:ablation-protocols}. \\
\bottomrule
\end{tabularx}
\end{table}

\clearpage
\begingroup
\subsection{Benchmark Metrics and Scoring}
\label{app:benchmark-scoring}

Table~\ref{tab:benchmark-scoring} summarizes the task-level metrics.
Implementation changes below are relative to the AgentScaling
v2.1.3 adapter. Finance-Agent and BrowseComp-Plus use Gemini 2.5
Flash-Lite with temperature 0; Plancraft and WorkBench use
deterministic environment or state checks.

\begin{table}[!htbp]
\centering
\caption{Task-level scoring across the four benchmarks.}
\label{tab:benchmark-scoring}
\begingroup
\MashTableSetup
\begin{tabularx}{\linewidth}{@{}lXX@{}}
\toprule
Benchmark & Task score & Scoring implementation \\
\midrule
Finance-Agent & Fraction of rubric items passed & Whole-checklist LLM grading \\
BrowseComp-Plus & Binary answer correctness & Reference-based LLM grading \\
Plancraft & Binary task success & Environment success flag \\
WorkBench & Binary expected-state success & Action replay and state comparison \\
\bottomrule
\end{tabularx}
\endgroup
\end{table}

\subsubsection{Finance-Agent}
For a task with $K$ rubric items, let $v_j\in\{0,1\}$ be the
Boolean verdict for item $j$ and $o_j$ its operator. The task score is
\[
s=\frac{1}{K}\sum_{j=1}^{K}p_j,\qquad
p_j=\begin{cases}
v_j, & o_j=\texttt{correctness},\\
1-v_j, & o_j=\texttt{contradiction}.
\end{cases}
\]
Thus, a correctness item passes on a true verdict, while a
contradiction item passes on a false verdict.

The modified scorer uses one whole-checklist request, following
the TacoMAS-style judge~\citep{xu2026tacomas}, in place of AgentScaling's per-criterion
requests. Each attempt sends the following single user message.
The field \texttt{\{safe\_answer\}} contains the candidate answer,
truncated to at most 12,000 characters;
\texttt{\{rubric\_text\}} contains the complete rubric serialized
as JSON; and \texttt{\{K\}} is the number of rubric items.
The task question and expected answer are not supplied as separate
fields.
\begin{verbatim}
You are an expert grader. Your task is to evaluate a candidate's
answer based on a specific rubric.

Candidate Answer:
"{safe_answer}"

Rubric (JSON checklist):
{rubric_text}

For each item in the rubric:
- If operator is "correctness": Does the answer support/contain
this fact? (True/False)
- If operator is "contradiction": Does the answer contradict this
fact? (True/False) NOTE: If answer contradicts, it is BAD.

Return only {"results":[true, false]} with exactly {K} JSON
booleans in rubric order.
Do not repeat operators or rubric text. Do not add explanations,
objects, keys, or code fences.
\end{verbatim}

After parsing and supported format normalization, the scorer
requires exactly one Boolean verdict per rubric item. Unresolved
grading failures are reported as scoring errors; an empty candidate
answer receives zero.

\clearpage
\subsubsection{BrowseComp-Plus}
The scorer reuses AgentScaling's scoring code and grader prompt,
with the grader model changed to Gemini 2.5 Flash-Lite.
The parsing rule is unchanged. The grader receives the question,
final response, and reference answer in a single user message:
\begin{verbatim}
Judge whether the following [response] to [question] is correct
or not based on the precise and unambiguous [correct_answer]
below.

[question]: {{question}}

[response]: {{response}}

extracted_final_answer: Extract the final exact answer from the
[response]. Put 'None' if there is no exact answer.

[correct_answer]: {{correct_answer}}

reasoning: Explain whether the extracted answer matches the
correct answer.

correct: Answer 'yes' if extracted_final_answer matches the
[correct_answer]. Answer 'no' otherwise.

confidence: The extracted confidence score from [response]. Put
100 if there is no confidence score.
\end{verbatim}

The task receives a score of 1 when the returned grader text
contains the case-sensitive substring \texttt{correct: yes},
and 0 otherwise. Extracted confidence is auxiliary metadata
and does not weight the score.

\subsubsection{Plancraft}
The adapter reuses AgentScaling's environment-based success check
over the Plancraft environment, with the success criterion unchanged.
A task succeeds when its target item is present in an inventory slot
other than slot 0 (the crafting output slot), or when the agent
invokes the impossible action for a task whose ground-truth
impossible flag is true. The explanation accompanying that action
is not graded by an LLM. The following pseudocode summarizes the
environment check; the adapter reads the final success flag.
\begin{verbatim}
if action is StopAction:  # invoked by the impossible tool
    success = task.impossible
else:
    observation = environment.step(action)
    success = any(
        item["type"] == task.target and slot != 0
        for slot, item in observation["inventory"].items()
    )

task_score = int(success)
\end{verbatim}

\clearpage
\subsubsection{WorkBench}
The local adapter implements WorkBench's outcome-centric evaluation,
replacing AgentScaling's structural action matching with state
comparison. It reads the environment's recorded action sequence.
Predicted and reference actions are replayed from the same initial
state, using the local replay and normalization rules. A task passes
when the resulting states match and the recorded execution-error
field is empty:
\begin{verbatim}
predicted_state = replay(predicted_actions, initial_state)
expected_state = replay(reference_actions, initial_state)

correct = (
    recorded_execution_error == ""
    and states_equal(predicted_state, expected_state)
)

task_score = int(correct)
\end{verbatim}

The comparison covers calendar, email, analytics,
project-management, and CRM state tables. Most string fields are
compared after lowercasing; \texttt{status}, \texttt{list\_name},
and \texttt{board} retain case sensitivity. The implementation
preserves row order. During replay, unparseable actions, unknown
tools, and actions that raise exceptions are skipped. The error
condition in the scoring rule checks the recorded execution-error
field.

Task performance uses \texttt{correct}. The separately computed
\texttt{exact\_match} field is diagnostic and is not used in the
task score.
\clearpage
\endgroup

\clearpage
\subsection{Baseline Adaptations and Search Budgets}
\label{app:baseline-details}

The comparison includes a static MAS reference, harness and context
optimizers, and MAS/workflow optimizers. We first describe their
editable components, then specify search budgets, stopping rules, and
compute accounting within the fixed MAS.

\paragraph{Common adaptation protocol.}
All methods follow the fixed configuration and partitions in
Appendix~\ref{app:experimental-setup}. Within those interfaces, each
retains its native feedback, search, and selection procedures over
prompts, state handling, coordination, aggregation, or control logic.

\subsubsection{Methods and Editable Components}
\paragraph{Static reference.}
\textbf{MAS-Centralized}~\citep{kim2026capable} instantiates the
centralized MAS architecture under our common backbone, tool, and
execution-budget configuration and serves as the unoptimized reference.
It retains the initial agent instructions, communication interfaces,
and orchestration logic without additional harness optimization.

\paragraph{Optimization methods.}
Table~\ref{tab:baseline-adaptations} summarizes the editable components
and native update mechanism of each optimizer. The adaptations use the
shared interfaces described above; their search schedules remain
method-specific.

\begin{table}[H]
\centering
\small
\setlength{\tabcolsep}{5pt}
\renewcommand{\arraystretch}{1.08}
\caption{Baseline adaptations within the fixed MAS. Each method edits
only behavior exposed by the existing interfaces; shared constraints
and Finance-Agent search schedules are specified in this section.}
\label{tab:baseline-adaptations}
\begin{tabularx}{\linewidth}{@{}>{\raggedright\arraybackslash}p{0.23\linewidth}>{\raggedright\arraybackslash}X@{}}
\toprule
Method & Editable components and update mechanism \\
\midrule
\multicolumn{2}{l}{\textit{Automatic harness and context optimization}} \\
GEPA\newline\citep{agrawal2026gepa} &
Natural-language instructions at agent and orchestration interfaces;
trajectory-based reflection, prompt revision, and Pareto-based
selection of complementary candidates. \\
\addlinespace
RHI\newline\citep{lee2026rhi} &
Textual agent instructions, communication contracts, and interaction
policies; revision from pairwise feedback accumulated over the
specification history. \\
\addlinespace
ACE\newline\citep{zhang2025ace} &
Context-interface playbook; generation, reflection, and incremental
curation of task strategies with fixed model weights. The playbook
is frozen after development. \\
\addlinespace
AHE\newline\citep{lin2026ahe} &
Harness descriptions, state or memory logic, and middleware, including
exposed tool behavior; evidence-driven editing and verification of
explicit, observable, and revertible components. \\
\addlinespace
Life-Harness\newline\citep{xu2026lifeharness} &
Environment contracts, procedural skills, action realization, and
trajectory regulation; runtime interventions for recurring failures
around a frozen agent through agent and orchestration interfaces. \\
\addlinespace
Meta-Harness\newline\citep{lee2026metaharness} &
Exposed harness-side code; outer-loop proposals informed by source code,
candidate scores, and development traces from earlier attempts. \\
\addlinespace
HarnessX\newline\citep{chen2026harnessx} &
Compositions of typed prompt, tool, memory, and control-flow primitives;
trace-driven substitution and recomposition compatible with the fixed
interfaces and topology. \\
\midrule
\multicolumn{2}{l}{\textit{MAS and workflow optimization}} \\
Meta-Team\newline\citep{hao2026metateam} &
Behavioral instructions, inter-agent coordination, and team execution
rules; experience aggregation and collaborative evolution with changes
realizable on the existing MAS graph. \\
\addlinespace
AFlow\newline\citep{zhang2024aflow} &
Code-represented workflows restricted to node instructions,
orchestration, scheduling, aggregation, and verification on the existing
MAS graph; Monte Carlo Tree Search with tree-structured experience,
code modification, and execution feedback. \\
\bottomrule
\end{tabularx}
\end{table}

\begingroup
\paragraph{Multi-agent prompt-optimization baselines.}
We include MASPO~\citep{wang2026maspo} and
MASPOB~\citep{hong2026maspob} to compare MASH with dedicated
multi-agent prompt optimization. Their adaptations follow the
common fixed-MAS protocol above: the agent roles and membership,
communication topology, task-execution backbone, tools, scorer,
data partitions, and per-task limits are shared with the other
baselines. Prompt optimization operates within that existing MAS
configuration, using each method's optimization and selection
procedure.

For both methods, we run the complete optimization independently
three times with different random seeds on the same data splits
across all four benchmarks. Each Finance-Agent run includes
independent optimization within each of the five folds.
The selected harnesses are frozen for three test repetitions under
the common evaluation protocol. Table~\ref{tab:main-results}
reports the mean and standard deviation across the three independent
optimization runs; Appendix~\ref{app:independent-runs} provides
the individual run scores.
\endgroup

\clearpage
\subsubsection{Search Budgets and Stopping Rules}
By default, each method receives 4,000 development MAS task attempts per
Finance-Agent fold, with no transfer of unused capacity between folds.
An attempt is one launched task execution, including initialization,
fresh parent and candidate runs, development validation, failures, and
whole-task retries. Reused results and reserved but unlaunched executions
incur no attempt; final testing is excluded. Search stops at the
schedule limit, the native stopping criterion, or when the remaining
budget cannot support continuation. Table~\ref{tab:baseline-schedules}
lists the schedules.
The round-sensitivity experiment uses an extended budget
(Appendix~\ref{app:hyperparameter-sensitivity}).

\begin{table}[H]
\centering
\small
\setlength{\tabcolsep}{4pt}
\renewcommand{\arraystretch}{1.08}
\caption{Finance-Agent search schedules per fold, subject to the common
4,000-task-attempt cap and method-specific stopping criteria.}
\label{tab:baseline-schedules}
\begin{tabular}{@{}p{0.18\linewidth}p{0.76\linewidth}@{}}
\toprule
Method & Search schedule \\
\midrule
GEPA &
Three-task reflection minibatches and eight-task validation;
budget-aware termination. \\

RHI &
At most 20 shared-harness revisions; 32 tasks once per revision;
parent results reused. \\

ACE &
Up to five passes over 32 tasks; at most five correction steps
per sample; per-sample curation. \\

AHE &
Ten evaluated artifacts: the initial version and nine revisions;
32 tasks with two repeats. \\

Life-Harness &
At most 20 new revisions beyond the initial artifact;
32 tasks with two repeats. \\

Meta-Harness &
At most 20 proposer sessions, with up to three initial candidates
each; 32 tasks with two repeats. \\

HarnessX &
At most 20 rounds and four candidate slots per round;
fresh parent evaluation; 32 tasks with two repeats. \\

Meta-Team &
One pass over 32 tasks; evolution after each original task;
at most one redo without retriggering evolution. \\

AFlow &
At most 20 tree expansions; 32 tasks with five repeats per artifact;
native convergence stopping. \\

MASH &
Twenty optimization rounds, including at most 19 generation rounds
with three initial candidate slots each; the final round
evaluates pending candidates. \\
\bottomrule
\end{tabular}
\end{table}

GEPA starts a mutation only when the remaining budget covers its
14-attempt evaluation path: three parent runs, three candidate runs,
and eight validation runs. HarnessX may stop before the round limit
with budget remaining. After development-set selection, each method
freezes its final artifact without an additional shared shortlist;
each test task is evaluated three times.

\subsubsection{Compute Accounting}
GPT-5.4 serves all optimizer-side roles in this Finance-Agent setup,
including reflection, diagnosis, generation, repair, internal judging,
and coding-agent subrequests. For ACE, the task-solving generator
belongs to the Gemini-based executor, while the reflector and curator
belong to the optimizer. Benchmark scoring is accounted for separately.

Development compute separates task-level MAS attempts from the
optimizer requests used to propose and assess changes. Attempts follow
the launch-based definition above; the request and token definitions
below are applied consistently across methods.

An optimizer call is an individual model request for analysis,
reflection, diagnosis, generation, repair, internal judging, or a
subsidiary optimization agent. Retries count whenever they issue a new
request, so one optimization session may contain several calls.
Tool invocations do not count as optimizer requests; restarting an entire
MAS task contributes another task attempt.

Input and output tokens are summed from request-level API usage.
Cached-input and reasoning-token fields are treated as breakdowns of
the corresponding totals rather than added again. MAS-internal model calls and
benchmark-grader usage are recorded separately from optimizer usage.

The measured development costs are reported in
Appendix~\ref{app:baseline-compute}.

\clearpage
\subsection{Shared Finance-Agent Search Protocol}
\label{app:ablation-protocols}

This protocol applies to centralized Finance-Agent comparisons of
diagnosis and candidate organization (Appendix~\ref{app:organization-protocol}),
complete optimization strategies (Appendix~\ref{app:strategy-resources}),
and filtering stages (Appendix~\ref{app:check-ablation-protocol}).
All use the same five-fold splits and initial harness
$H_0$, allow at most three initial candidates per generation round,
and follow the round sequence and evaluations below. Each configuration
runs its own search under the shared MAS setup in
Appendix~\ref{app:experimental-setup}.

\paragraph{Round sequence.}
Each fold uses 20 rounds, with different roles for the first,
intermediate, and final rounds:
\begin{itemize}[leftmargin=*,itemsep=0.3em]
    \item \textbf{Round 1: initialize.} Execute $H_0$ on the 32
    diagnosis tasks, diagnose the trajectories, and generate and filter
    the first candidate batch. The parent is unchanged.
    \item \textbf{Rounds 2--19: evaluate and generate.} Compare the
    retained candidates with a fixed parent and select at most one
    update. Diagnose the collected trajectories, then generate and
    filter the next batch against the resulting harness.
    \item \textbf{Round 20: evaluate and freeze.} Evaluate the pending
    candidates and freeze the resulting harness, without generating
    another batch.
\end{itemize}
Each fold therefore has 19 generation opportunities and at most
19 parent updates.

\paragraph{Filtering and adoption.}
The checks follow Appendix~\ref{app:candidate-protocol}; Stage 1
requires no MAS execution. Table~\ref{tab:finance-evaluation-settings}
collects the execution-based comparisons used in these analyses.
The screening subset is fixed within each fold, and the validation
tasks are disjoint from diagnosis tasks.

\begin{table}[H]
\centering
\small
\setlength{\tabcolsep}{4pt}
\renewcommand{\arraystretch}{1.08}
\caption{Finance-Agent evaluation settings for the shared search
protocol. Repetitions apply to each harness on each task; parent and
candidate both execute. Thresholds use the 0--100 score scale:
0.5 and 0.25 points correspond to 0.005 and 0.0025 on the $[0,1]$ scale.}
\label{tab:finance-evaluation-settings}
\begin{tabularx}{\linewidth}{@{}>{\raggedright\arraybackslash}p{0.21\linewidth}>{\raggedright\arraybackslash}p{0.19\linewidth}c>{\raggedright\arraybackslash}X@{}}
\toprule
Comparison & Task set & Repetitions & Decision \\
\midrule
Stage 2: effectiveness & 4 screening tasks & 1 &
Advance if mean gain is at least 0.5 points. \\
Stage 3: robustness & 8 validation tasks & 1 &
Retain if mean gain is at least 0.25 points. \\
Next-round adoption & 32 diagnosis tasks & 2 &
Adopt the completed candidate with the largest strictly positive
mean gain; otherwise retain the parent. \\
\bottomrule
\end{tabularx}
\end{table}

Candidates passing Stage 3 enter the next round's adoption comparison.
This comparison is separate from the three filtering stages. Parent
scores are not reused across candidates, stages, or rounds. The
adoption criterion also differs from the $>1$-point threshold used
to report candidate improvement rates in
Appendix~\ref{app:candidate-quality-cost}.

\begingroup
\subsection{Code and Experimental Artifacts}
\label{app:artifact-index}

The \href{https://github.com/GeniusNew/MASH_code_public/blob/main/README.md}
{public repository} provides the MASH core implementation,
including the search controller, execution tracing, hook runtime,
optimizer integration, and artifact validation utilities.
\clearpage
\endgroup

\clearpage
\begingroup
\section{Results Across Independent Optimization Runs}
\label{app:independent-runs}

We provide the individual optimization-run results underlying
Tables~\ref{tab:main-results} and~\ref{tab:representation-ablation}.
Each run repeats the complete optimization with a different random seed
on the same data splits. The selected harness is then frozen for three
test repetitions. A run score averages these test repetitions and, for
Finance-Agent, pools the five held-out folds after independent
optimization within each fold. The three optimization runs, rather than
the frozen-harness test repetitions, are the units used for the
between-run statistics. All scores below are multiplied by 100.

\paragraph{Main-comparison runs.}
Tables~\ref{tab:runs-finance}--\ref{tab:runs-workbench} list the three
run scores for every optimized method. Run 1 denotes the first
independent optimization run; the mean and standard deviation across
runs are reported in Table~\ref{tab:main-results}.

\begin{table}[htbp]
\centering
\caption{Finance-Agent: test scores across three independent optimization runs under the centralized workflow.}
\label{tab:runs-finance}
\begingroup
\MashTableSetup
\begin{tabular*}{\linewidth}{@{\extracolsep{\fill}}lrrr@{}}
\toprule
Method & Run 1 & Run 2 & Run 3 \\
\midrule
GEPA & 39.32 & 38.27 & 41.03 \\
RHI & 43.02 & 42.48 & 44.23 \\
ACE & 38.23 & 37.75 & 36.86 \\
AHE & 46.83 & 45.46 & 47.12 \\
Life-Harness & 43.49 & 44.72 & 41.95 \\
Meta-Harness & 47.24 & 48.75 & 49.37 \\
HarnessX & 48.76 & 47.47 & 49.85 \\
Meta-Team & 41.86 & 42.27 & 43.63 \\
AFlow & 42.85 & 41.72 & 42.16 \\
MASPO & 45.12 & 46.30 & 43.93 \\
MASPOB & 43.28 & 41.96 & 44.25 \\
\textbf{MASH} & \textbf{52.88} & \textbf{51.11} & \textbf{52.13} \\
\bottomrule
\end{tabular*}
\endgroup
\end{table}

\begin{table}[htbp]
\centering
\caption{BrowseComp-Plus: test scores across three independent optimization runs under the centralized workflow.}
\label{tab:runs-browse}
\begingroup
\MashTableSetup
\begin{tabular*}{\linewidth}{@{\extracolsep{\fill}}lrrr@{}}
\toprule
Method & Run 1 & Run 2 & Run 3 \\
\midrule
GEPA & 21.00 & 20.00 & 19.33 \\
RHI & 23.00 & 23.67 & 24.33 \\
ACE & 20.33 & 19.67 & 19.00 \\
AHE & 24.00 & 23.33 & 25.67 \\
Life-Harness & 21.67 & 22.33 & 20.67 \\
Meta-Harness & 25.00 & 26.67 & 24.33 \\
HarnessX & 25.33 & 27.67 & 26.67 \\
Meta-Team & 21.67 & 20.67 & 21.33 \\
AFlow & 22.33 & 21.67 & 24.00 \\
MASPO & 23.33 & 24.67 & 22.67 \\
MASPOB & 21.67 & 22.67 & 24.00 \\
\textbf{MASH} & \textbf{29.00} & \textbf{28.67} & \textbf{30.33} \\
\bottomrule
\end{tabular*}
\endgroup
\end{table}

\begin{table}[htbp]
\centering
\caption{Plancraft: test scores across three independent optimization runs under the centralized workflow.}
\label{tab:runs-plancraft}
\begingroup
\MashTableSetup
\begin{tabular*}{\linewidth}{@{\extracolsep{\fill}}lrrr@{}}
\toprule
Method & Run 1 & Run 2 & Run 3 \\
\midrule
GEPA & 51.67 & 51.33 & 52.33 \\
RHI & 54.33 & 55.00 & 55.33 \\
ACE & 51.00 & 49.67 & 50.33 \\
AHE & 55.33 & 56.67 & 54.67 \\
Life-Harness & 54.00 & 52.67 & 55.00 \\
Meta-Harness & 56.33 & 58.00 & 56.67 \\
HarnessX & 56.00 & 54.33 & 55.67 \\
Meta-Team & 55.00 & 54.33 & 53.67 \\
AFlow & 57.00 & 56.33 & 55.67 \\
MASPO & 56.67 & 57.67 & 55.33 \\
MASPOB & 55.00 & 56.00 & 54.33 \\
\textbf{MASH} & \textbf{61.67} & \textbf{59.00} & \textbf{62.00} \\
\bottomrule
\end{tabular*}
\endgroup
\end{table}

\begin{table}[htbp]
\centering
\caption{WorkBench: test scores across three independent optimization runs under the centralized workflow.}
\label{tab:runs-workbench}
\begingroup
\MashTableSetup
\begin{tabular*}{\linewidth}{@{\extracolsep{\fill}}lrrr@{}}
\toprule
Method & Run 1 & Run 2 & Run 3 \\
\midrule
GEPA & 34.67 & 35.00 & 33.33 \\
RHI & 38.00 & 37.33 & 40.33 \\
ACE & 33.67 & 34.00 & 35.33 \\
AHE & 40.00 & 42.00 & 39.67 \\
Life-Harness & 42.00 & 40.67 & 43.33 \\
Meta-Harness & 42.33 & 44.67 & 43.67 \\
HarnessX & 43.33 & 41.67 & 45.33 \\
Meta-Team & 40.67 & 39.33 & 41.33 \\
AFlow & 39.00 & 39.67 & 40.67 \\
MASPO & 40.67 & 43.33 & 41.33 \\
MASPOB & 39.67 & 38.33 & 41.33 \\
\textbf{MASH} & \textbf{48.00} & \textbf{49.00} & \textbf{47.67} \\
\bottomrule
\end{tabular*}
\endgroup
\end{table}

\paragraph{Finance-Agent fold results.}
Table~\ref{tab:finance-run-folds} gives the five fold scores for
Meta-Harness, HarnessX, and MASH in each optimization run.
Each fold score averages the frozen-harness test repetitions on that
fold; the final column averages the five folds.

\begin{table}[htbp]
\centering
\caption{Finance-Agent test scores by fold and optimization run under the centralized workflow.}
\label{tab:finance-run-folds}
\begingroup
\MashTableSetup
\begin{tabular*}{\linewidth}{@{\extracolsep{\fill}}lrrrrrrr@{}}
\toprule
Method & Run & Fold 1 & Fold 2 & Fold 3 & Fold 4 & Fold 5 & Mean \\
\midrule
Meta-Harness & 1 & 40.25 & 53.70 & 43.00 & 57.50 & 41.75 & 47.24 \\
 & 2 & 40.80 & 55.50 & 43.35 & 60.60 & 43.50 & 48.75 \\
 & 3 & 40.40 & 53.80 & 48.30 & 58.65 & 45.70 & 49.37 \\
\midrule
HarnessX & 1 & 40.25 & 54.95 & 44.20 & 61.00 & 43.40 & 48.76 \\
 & 2 & 37.10 & 53.65 & 46.95 & 56.25 & 43.40 & 47.47 \\
 & 3 & 39.60 & 56.20 & 50.10 & 59.80 & 43.55 & 49.85 \\
\midrule
MASH & 1 & 43.65 & 60.25 & 47.75 & 65.85 & 46.90 & 52.88 \\
 & 2 & 42.15 & 60.15 & 49.20 & 62.00 & 42.05 & 51.11 \\
 & 3 & 44.65 & 59.10 & 48.20 & 64.20 & 44.50 & 52.13 \\
\bottomrule
\end{tabular*}
\endgroup
\end{table}

\paragraph{Uncertainty of the mean score differences.}
Table~\ref{tab:run-gain-intervals} reports MASH's score differences
relative to HarnessX and Meta-Harness. We compute two-sided 95\%
Welch $t$ confidence intervals for the difference of means, using the
three independent optimization-run scores per method, unequal sample
variances, and Welch--Satterthwaite degrees of freedom.
The gains and interval endpoints in this table are calculated from the
individual run scores listed above, before rounding the final results
to two decimal places. Each interval is pointwise for its comparison.
All eight confidence intervals lie above zero.

\begin{table}[htbp]
\centering
\caption{MASH minus each baseline in score points. Run 1 gain uses the first optimization run; mean gain uses all three optimization runs. Intervals are 95\% Welch confidence intervals.}
\label{tab:run-gain-intervals}
\begingroup
\MashTableSetup
\begin{tabular*}{\linewidth}{@{\extracolsep{\fill}}llrrr@{}}
\toprule
Benchmark & Baseline & Run 1 gain & Mean gain & 95\% CI \\
\midrule
Finance-Agent & HarnessX & $+4.12$ & $+3.35$ & $[+0.89, +5.81]$ \\
 & Meta-Harness & $+5.64$ & $+3.59$ & $[+1.29, +5.89]$ \\
\midrule
BrowseComp-Plus & HarnessX & $+3.67$ & $+2.78$ & $[+0.35, +5.20]$ \\
 & Meta-Harness & $+4.00$ & $+4.00$ & $[+1.52, +6.48]$ \\
\midrule
Plancraft & HarnessX & $+5.67$ & $+5.56$ & $[+2.17, +8.95]$ \\
 & Meta-Harness & $+5.34$ & $+3.89$ & $[+0.50, +7.28]$ \\
\midrule
WorkBench & HarnessX & $+4.67$ & $+4.78$ & $[+0.80, +8.76]$ \\
 & Meta-Harness & $+5.67$ & $+4.67$ & $[+2.26, +7.07]$ \\
\bottomrule
\end{tabular*}
\endgroup
\end{table}

\paragraph{Component-ablation runs.}
Table~\ref{tab:component-ablation-runs} lists the three optimization-run
scores for the four Finance-Agent component variants, following
Appendix~\ref{app:organization-protocol}. Run 1 denotes the first optimization run.
All variants retain node-, edge-, and graph-level targets; disabling
cross-level generation restricts each generated constraint to one level.
Table~\ref{tab:representation-ablation} reports the corresponding
means and standard deviations. Combining both components gives the
highest mean, with a 0.83-point mean gain from cross-level generation
when structured diagnosis is already present.

\begin{table}[htbp]
\centering
\caption{Finance-Agent component ablation across three independent optimization runs. Scores average the five folds and frozen-harness test repetitions.}
\label{tab:component-ablation-runs}
\begingroup
\MashTableSetup
\begin{tabular*}{\linewidth}{@{\extracolsep{\fill}}llrrr@{}}
\toprule
\shortstack[l]{Structured\\diagnosis} & \shortstack[l]{Cross-level\\generation} & Run 1 & Run 2 & Run 3 \\
\midrule
No & No & 48.60 & 47.18 & 49.02 \\
No & Yes & 50.57 & 49.04 & 51.41 \\
Yes & No & 51.44 & 52.14 & 50.06 \\
Yes & Yes & 52.88 & 51.11 & 52.13 \\
\bottomrule
\end{tabular*}
\endgroup
\end{table}

\clearpage
\endgroup

\clearpage
\section{Cross-Level Optimization Analyses}
\label{app:results}
We first test which levels should remain editable and measure diagnosis
redundancy on fixed initial-harness trajectories. We then examine candidate
quality and accepted revisions under different organization strategies.
Finally, we compare observable failures in frozen test executions.

\subsection{Editable-Level Ablation}
\label{app:level-ablation}

We test which levels should remain editable after improving local
execution. All variants start from $H_N$, obtained after 20 node-only
rounds, and continue for 20 rounds: A edits nodes, B edits edges and
graphs with node rules frozen, and C edits all three levels. Frozen
rules remain active. All use GPT-5.4, identical evaluation rules, at
most three initial candidates per round, and session caps of 11 for
analysis and 13 for generation.

\begin{table}[H]
\centering
\small
\setlength{\tabcolsep}{4.0pt}
\renewcommand{\arraystretch}{1.08}
\caption{
Editable-level ablation: 20 additional rounds from the same
node-strengthened $H_N$, following 20 node-only rounds.
Scores are multiplied by 100; best scores are bold.
Improvement is C minus A in score points.
}
\label{tab:ablation-admissible-levels}
\begin{tabular}{lcccc}
\toprule
Variant & Finance-Agent & BrowseComp-Plus & Plancraft & WorkBench \\
\midrule
A: Node & 48.12 & 25.00 & 61.00 & 45.00 \\
B: Edge + Graph & 51.60 & 28.00 & 58.33 & 44.33 \\
C: Node + Edge + Graph & \textbf{53.12} & \textbf{29.33} & \textbf{61.33} & \textbf{47.67} \\
\midrule
\quad Improvement: C $-$ A & +5.00 & +4.33 & +0.33 & +2.67 \\
\bottomrule
\end{tabular}
\end{table}

Keeping all three levels editable yields the highest score on every
benchmark, with larger gains over node-only continuation on
Finance-Agent and BrowseComp-Plus than on Plancraft. Coordination-only
continuation helps on the first two benchmarks but trails node-only
continuation on Plancraft and WorkBench, supporting joint optimization
of local and coordination decisions.

\begingroup
\paragraph{Node-only and full MASH across communication workflows.}
We also compare node-only and full MASH on Finance-Agent under
the centralized, decentralized, and hybrid workflows.
For each workflow and fold, both conditions start independently
from the same initial harness $H_0$ and follow the main experiment's
schedule of at most 20 optimization rounds
(Appendix~\ref{app:ablation-protocols}).
The node-only condition permits node-level harness edits only;
full MASH permits edits at all three levels.
The optimizer, structured diagnosis, filtering pipeline, and
budget are held fixed within this comparison.
Table~\ref{tab:workflow-node-full} reports these workflow-level
results alongside the initial MAS scores.
Full MASH improves over node-only by 6.06, 2.22, and 3.37 points
on centralized, decentralized, and hybrid workflows, respectively.
These comparisons measure the contribution of including edge-
and graph-level harnesses in workflows with inter-agent messages.

\begin{table}[!htbp]
\centering
\caption{Node-only and full MASH on Finance-Agent across three
communication workflows, starting from $H_0$ with at most
20 optimization rounds per condition. Test scores are multiplied by 100;
$\Delta$ is full MASH minus node-only.}
\label{tab:workflow-node-full}
\begingroup
\MashTableSetup
\begin{tabular*}{\linewidth}{@{\extracolsep{\fill}}lrrrr@{}}
\toprule
Workflow & Initial MAS & Node-only & Full MASH & $\Delta$ \\
\midrule
Centralized & 39.97 & 46.82 & \textbf{52.88} & +6.06 \\
Decentralized & 38.26 & 46.75 & \textbf{48.97} & +2.22 \\
Hybrid & 39.55 & 47.39 & \textbf{50.76} & +3.37 \\
\bottomrule
\end{tabular*}
\endgroup
\end{table}
\endgroup

\appendixneedspace{0.48\textheight}
\subsection{Diagnosis Redundancy}
\label{app:diagnosis-redundancy}

On centralized Finance-Agent, MASH (F) and free-form diagnosis (O)
use the same fixed $H_0$ trajectories in each of five folds, with matched
models, raw-evidence access, and analysis allowances. GPT-5.4 groups each
method's final diagnosis records in a separate evaluation context. It uses
a common prompt without receiving the methods' consolidation assignments.
Trajectory-supported records sharing an originating defect, including
propagated symptoms, are grouped within each execution; independent
defects and instances in different executions remain separate.

A program aggregates groups across folds. For each method,
$N_{\mathrm{diag}}$ counts supported records before grouping;
$K_{\mathrm{rc}}$ counts the distinct execution-specific root-cause instances
represented in those records. The $N_{\mathrm{diag}}-K_{\mathrm{rc}}$ repeated records give:
\[
R_{\mathrm{red}}
=\frac{N_{\mathrm{diag}}-K_{\mathrm{rc}}}{N_{\mathrm{diag}}}
\times100\%.
\]

\begin{table}[H]
\centering
\small
\setlength{\tabcolsep}{8pt}
\renewcommand{\arraystretch}{1.08}
\caption{Diagnosis redundancy on shared $H_0$ trajectories across
five folds of centralized Finance-Agent.}
\label{tab:diagnosis-redundancy}
\begin{tabular}{@{}lrr@{}}
\toprule
Metric & Free-form (O) & MASH (F) \\
\midrule
Supported diagnoses $N_{\mathrm{diag}}$ & 536 & 440 \\
Distinct root-cause instances $K_{\mathrm{rc}}$ & 329 & 340 \\
Repeated diagnoses $N_{\mathrm{diag}}-K_{\mathrm{rc}}$ & 207 & 100 \\
Redundancy rate $R_{\mathrm{red}}$ & 38.62\% & 22.73\% \\
\bottomrule
\end{tabular}
\end{table}

MASH reduces diagnosis redundancy by 15.89 percentage points
(Figure~\ref{fig:candidate-quality-cost}(c)), with fewer supported
diagnoses and more distinct root-cause instances.

\begingroup
\subsection{Search-Time Diagnosis Audit}
\label{app:search-diagnosis-audit}

We audit diagnoses produced during optimization on Finance-Agent
under the centralized workflow across five folds, with 19 diagnosis
rounds per fold (95 diagnosis rounds in total).
This analysis examines search-time failure consolidation and diagnosis,
separately from the frozen-harness test-trajectory labeling in
Appendix~\ref{app:failure-labeling-protocol}.
The audit uses a custom assessment pipeline with GPT-5.5 as the
LLM judge. Human review evaluates a randomly sampled subset of
merge groups.

\paragraph{Incident population and failure types.}
The audit ledger contains 2,156 observed failure incidents: raw
failure events observed during search. An incident is the unit of
the applicability and diagnosis analysis; merge groups and failure
pairs are used for consolidation analysis.
We classify incidents as pure information-gap (necessary information
is missing or inadequately transmitted), information-sufficient
(necessary information is available, but reasoning, calculation, or
source selection fails), mixed (both an information gap and a
reasoning, calculation, or information-use error), or unknown
(insufficient evidence for reliable classification).
The first three categories are mutually exclusive and exhaust the
2,083 assessable incidents; the remaining 73 incidents are unknown.
These operational audit categories describe a different dimension
from the node/edge/graph responsibility levels.
Table~\ref{tab:search-audit-incidents} gives the incident counts.

\begin{table}[!htbp]
\centering
\caption{Observed failure incidents and correctly diagnosed
incidents in the search-time audit.}
\label{tab:search-audit-incidents}
\begingroup
\MashTableSetup
\begin{tabular*}{\linewidth}{@{\extracolsep{\fill}}lrr@{}}
\toprule
Failure type & Observed & Correctly diagnosed \\
\midrule
Pure information-gap & 1,487 & 1,233 \\
Information-sufficient & 455 & 308 \\
Mixed & 141 & 82 \\
Unknown & 73 & --- \\
\midrule
Total & 2,156 & 1,623 \\
\bottomrule
\end{tabular*}
\endgroup
\end{table}

\paragraph{Failure consolidation.}
The group-level ledger contains 506 multi-failure merge groups:
455 correct, 41 incorrect, and 10 uncertain. Excluding the uncertain
groups gives 496 assessable groups, of which 91.7\% are correct
(Table~\ref{tab:search-diagnosis-audit-results}). A group is correct
when the shared root cause of all its members is supported by
evidence; any incorrectly merged member pair prevents the group
from being counted as correct.

At the pair level, the reference judgments identify 935 failure
pairs with a shared root cause: 822 true positives and 113 false
negatives. The 113 pairs were not correctly merged by MASH, giving
a missed-shared-cause rate of $113/(822+113)=12.1\%$.
The ledger also records 66 false positives, so the number of
assessable predicted-positive pairs is $822+66=888$.

\paragraph{Human review of merge groups.}
The random review sample contains 155 merge groups, with
LLM--human agreement on 143 groups (92.3\%) and human--human
agreement on 149 groups (96.1\%). Agreement compares judgments
on the same merge group. Table~\ref{tab:search-audit-human-review}
reports the fold-level counts. Including additional difficult
cases, the human-review ledger contains 190 unique merge groups.

\begin{table}[!htbp]
\centering
\caption{Random human review of merge groups by fold.}
\label{tab:search-audit-human-review}
\begingroup
\MashTableSetup
\begin{tabular*}{\linewidth}{@{\extracolsep{\fill}}lrr@{}}
\toprule
Fold & Reviewed groups & LLM--human agreements \\
\midrule
1 & 26 & 24 \\
2 & 35 & 33 \\
3 & 32 & 30 \\
4 & 32 & 29 \\
5 & 30 & 27 \\
\midrule
Total & 155 & 143 \\
\bottomrule
\end{tabular*}
\endgroup
\end{table}

\paragraph{Information-gap applicability and diagnosis.}
Of the 2,083 assessable incidents, 1,487 (71.4\%) are pure
information-gap failures. Including mixed failures, 1,628 incidents
(78.2\%) involve an information gap
(Table~\ref{tab:search-diagnosis-audit-results}).
Within the 1,487 pure information-gap cases, 1,233 diagnoses are
correct (82.9\%). Applicability describes the type of failure;
diagnostic correctness describes how well MASH identifies it.

\paragraph{Overall structured diagnosis coverage.}
The ledger records correct structured diagnoses for 1,233 pure-gap,
308 information-sufficient, and 82 mixed incidents. Coverage over
the assessable population is therefore
\[
\mathrm{Coverage}_{\mathrm{assessable}}
=\frac{1,233+308+82}{1,487+455+141}
=\frac{1,623}{2,083}=77.9\%.
\]
Using all observed incidents, including the 73 unknown cases,
gives $\mathrm{Coverage}_{\mathrm{all}}=1,623/2,156=75.3\%$.

\paragraph{Errors with sufficient information.}
Reasoning or calculation can fail even when the necessary information
is available. Such failures are recorded as node-level errors with
their diagnostic reason. The diagnosis prompt permits an empty
missing-information field and instructs the model not to mark
information as missing merely because a later agent failed to use it.
When consolidating these cases, the diagnosis retains the reasoning
or calculation error rather than requiring an information-gap label.

Table~\ref{tab:search-diagnosis-audit-results} reports the outcomes for 455
information-sufficient failures. MASH diagnoses 308 correctly (67.7\%),
incorrectly attributes 49 to information gaps (10.8\%), and misses
or provides unsupported diagnoses for 98 (21.5\%).

\begin{table}[!htbp]
\centering
\caption{Search-time diagnosis audit on Finance-Agent under the
centralized workflow, across five folds. Consolidation correctness
and human agreement use merge groups; missed shared causes use
reference-positive pairs; applicability and diagnosis use incidents.
Correct information-gap diagnosis is evaluated
within pure information-gap cases; the three non-gap outcomes
partition the 455 information-sufficient cases.}
\label{tab:search-diagnosis-audit-results}
\begingroup
\MashTableSetup
\begin{tabular*}{\linewidth}{@{\extracolsep{\fill}}lrr@{}}
\toprule
Metric & Count & Rate \\
\midrule
\multicolumn{3}{l}{\textbf{Failure consolidation}} \\
Correct merge groups & 455/496 & 91.7\% \\
Incorrect merge groups & 41/496 & 8.3\% \\
Missed shared-cause pairs & 113/935 & 12.1\% \\
LLM--human group agreement & 143/155 & 92.3\% \\
Human--human group agreement & 149/155 & 96.1\% \\
\midrule
\multicolumn{3}{l}{\textbf{Information-gap applicability and diagnosis}} \\
Pure information-gap applicability & 1,487/2,083 & 71.4\% \\
Information-gap applicability (including mixed) & 1,628/2,083 & 78.2\% \\
Correct information-gap diagnosis & 1,233/1,487 & 82.9\% \\
Structured diagnosis coverage (assessable) & 1,623/2,083 & 77.9\% \\
Structured diagnosis coverage (all observed) & 1,623/2,156 & 75.3\% \\
\midrule
\multicolumn{3}{l}{\textbf{Information-sufficient errors}} \\
Correct non-gap diagnosis & 308/455 & 67.7\% \\
False information-gap attribution & 49/455 & 10.8\% \\
Missed / unsupported diagnosis & 98/455 & 21.5\% \\
\bottomrule
\end{tabular*}
\endgroup
\end{table}
\clearpage
\endgroup

\appendixneedspace{0.34\textheight}
\subsection{Diagnosis, Candidate Quality, and Accepted Revisions}
\label{app:organization-protocol}

F/U/O/R vary diagnosis format (structured or free-form) and generation
protocol (cross-level or single-level). L is a separate level-wise search
strategy (Table~\ref{tab:organization-variants}).
Variants share the schedule in Appendix~\ref{app:ablation-protocols},
optimizer-session limits, and evidence permissions, and each diagnoses
its own traces. Structured diagnosis records required, received, and
missing information.

\begin{table}[H]
\centering
\small
\setlength{\tabcolsep}{5pt}
\renewcommand{\arraystretch}{1.08}
\caption{Organization of failure diagnosis and candidates in the
F/U/O/R organization comparison and the complete F/R/L strategy comparison.}
\label{tab:organization-variants}
\begin{tabularx}{\textwidth}{@{}lXX@{}}
\toprule
Variant & Failure diagnosis & Candidate composition \\
\midrule
F (MASH) & Structured &
Shared repair hypothesis; targets may span node, edge, and graph. \\
U & Structured (as F) & Single-level; multiple targets allowed. \\
O & Free-form & Cross-level generation protocol as F. \\
R & Free-form & Single-level; multiple targets allowed. \\
L & Separate analysis by level & Multiple targets within one level;
no candidate is required for every level. \\
\bottomrule
\end{tabularx}
\end{table}

All three levels remain available throughout F/U/O/R search; each
candidate may revise multiple hooks within its permitted level scope.
All variants share the filtering-stage sequence and adoption procedure.
Stage 1 checks interface validity, implementation of declared edits
(proposal consistency), and behavior cases. F/O additionally check
consistency with their frozen repair plans as part of the cross-level
generation protocol; U/R have no such requirement. In the F/R/L
strategy comparison, all variants share basic implementation checks,
and only F applies plan-specific consistency checks
(Appendix~\ref{app:candidate-protocol}).

For
Table~\ref{tab:representation-ablation}, we repeat the complete
optimization of each variant three times with different random seeds
on the same five-fold splits. Each optimization run uses the frozen-harness
test evaluation described in Section~\ref{sec:exp-settings}; the table
reports the mean and standard deviation across these three optimization runs.

\appendixneedspace{0.38\textheight}
\subsubsection{Candidate Quality}
\label{app:candidate-quality-cost}

Table~\ref{tab:constraint-quality} tracks filtering, paired evaluation,
and adoption across five Finance-Agent folds.

\emph{Submission and evaluation counts.} The sequence $N\to Q\to M$
tracks submissions, filtering survivors, and completed paired comparisons.
$N$ counts each initial implementation or resubmission after legality repair
or quality revision once. $Q$ counts versions passing all three checks.
$M$ counts completed next-round comparisons on 32 diagnosis tasks, with two
repetitions per task per harness. Unfinished comparisons $Q-M$ number
9, 10, 8, 10, and 10 for F/U/O/R/L, respectively.

\emph{Improvement and adoption counts.} $I$ and $A$ apply different criteria
to completed comparisons, not successive filters. $I$ counts candidates with paired gain $\bar\Delta_{\mathrm{dis}}>0.01$ (over one point on the 100-point scale).
$A$ counts accepted revisions: at most one per round, with the largest strictly
positive gain; adoption need not meet the reporting threshold.
$I/M$ uses completed comparisons as its denominator; $A/N$ uses all submissions.

\begin{table}[H]
\centering
\small
\setlength{\tabcolsep}{4.0pt}
\renewcommand{\arraystretch}{1.08}
\caption{
Candidate quality on centralized Finance-Agent, pooled across five folds.
Entries give numerator/denominator (percentage): $Q/N$ measures
filtering passage, $I/M$ improvement among completed comparisons,
and $A/N$ adoption among submissions. Column maxima are bold. Variants follow
Table~\ref{tab:organization-variants}.
}
\label{tab:constraint-quality}
\begin{tabular}{@{}lccc@{}}
\toprule
Variant
& \shortstack{Filtering pass rate\\$Q/N$}
& \shortstack{Improvement rate\\$I/M$}
& \shortstack{Accepted revision rate\\$A/N$} \\
\midrule
F (MASH)
& 96/164 (\textbf{58.5\%})
& 56/87 (\textbf{64.4\%})
& 29/164 (17.7\%) \\
U
& 94/173 (54.3\%)
& 51/84 (60.7\%)
& 38/173 (\textbf{22.0\%}) \\
O
& 86/169 (50.9\%)
& 43/78 (55.1\%)
& 30/169 (17.8\%) \\
R
& 84/181 (46.4\%)
& 36/74 (48.6\%)
& 29/181 (16.0\%) \\
L
& 94/176 (53.4\%)
& 49/84 (58.3\%)
& 33/176 (18.8\%) \\
\bottomrule
\end{tabular}
\end{table}

F (MASH) has the highest filtering pass and improvement rates, whereas
U has the highest adoption rate. F also achieves the highest final test
score (Table~\ref{tab:representation-ablation} and
Appendix~\ref{app:strategy-resources}).

\appendixneedspace{0.32\textheight}
\subsubsection{Accepted Revision Distribution}
\label{app:accepted-revision-distribution}

We next examine which targets change when MASH adopts a candidate.
The 29 accepted revisions in
Table~\ref{tab:constraint-quality} contain 64 hook modifications across
five Finance-Agent folds, counting each modified hook once per update.
Table~\ref{tab:accepted-revision-distribution} reports both the hook
levels and mutually exclusive revision scopes.

\begin{table}[H]
\centering
\small
\setlength{\tabcolsep}{5pt}
\renewcommand{\arraystretch}{1.08}
\caption{
Level distribution of accepted MASH revisions across the five
Finance-Agent folds. Hook-level shares are computed over 64 hook
modifications; revision-level shares are computed over 29
accepted revisions. Cross-level revisions modify two or more levels.
}
\label{tab:accepted-revision-distribution}
\begin{tabular}{lrr}
\toprule
\multicolumn{3}{l}{\textbf{Hook-level distribution}} \\
Level & Hook modifications & Share \\
\midrule
Node  & 24 & 37.5\% \\
Edge  & 15 & 23.4\% \\
Graph & 25 & 39.1\% \\
\midrule
\multicolumn{3}{l}{\textbf{Revision-level distribution}} \\
Scope & Accepted revisions & Share \\
\midrule
Node only   & 9  & 31.0\% \\
Edge only   & 3  & 10.3\% \\
Graph only  & 5  & 17.2\% \\
Cross-level & 12 & 41.4\% \\
\bottomrule
\end{tabular}
\end{table}

Of 29 accepted revisions, 20 involve edge or graph targets and 12
span levels, the largest revision category.

\begingroup
\paragraph{Accepted hook modifications across workflows.}
Table~\ref{tab:accepted-hooks-workflows} extends the hook-level
counts to all four Finance-Agent workflows. Each modified hook
is counted once per accepted revision, and shares use the total
hook modifications within the corresponding workflow as the
denominator. One accepted revision can modify multiple hooks.
Edge and graph hooks together account for 40/64 (62.5\%) of
modifications in centralized, 39/67 (58.2\%) in decentralized,
and 45/68 (66.2\%) in hybrid workflows.

The independent workflow has no intermediate inter-agent messages
and interacts through terminal aggregation. Its accepted changes
comprise 37 node-hook and 10 graph-hook modifications, with no
edge-hook modifications. Graph hooks account for 21.3\% of its
47 hook modifications and govern terminal aggregation.

\begin{table}[!htbp]
\centering
\caption{Accepted revisions and hook modifications across
Finance-Agent workflows. Node, edge, and graph cells report
counts with shares of that workflow's total hook modifications.
The revisions column counts accepted parent-harness updates.}
\label{tab:accepted-hooks-workflows}
\begingroup
\MashTableSetup
\begin{tabular*}{\linewidth}{@{\extracolsep{\fill}}lrrrrr@{}}
\toprule
Workflow & Revisions & Node & Edge & Graph & Total \\
\midrule
Centralized & 29 & 24 (37.5\%) & 15 (23.4\%) & 25 (39.1\%) & 64 \\
Decentralized & 27 & 28 (41.8\%) & 26 (38.8\%) & 13 (19.4\%) & 67 \\
Hybrid & 30 & 23 (33.8\%) & 18 (26.5\%) & 27 (39.7\%) & 68 \\
Independent & 23 & 37 (78.7\%) & 0 (0.0\%) & 10 (21.3\%) & 47 \\
\bottomrule
\end{tabular*}
\endgroup
\end{table}
\endgroup

\appendixneedspace{0.32\textheight}
\subsection{Failures Across Node, Edge, and Graph}
\label{app:collaboration-errors}

We measure failures post hoc in frozen-harness test executions under the
centralized workflow, independently of the diagnoses used during optimization.
Appendix~\ref{app:failure-labeling-protocol} defines the potentially
overlapping node, edge, and graph labels for non-perfect executions.

\subsubsection{Finance-Agent Breakdown}
Each method has 150 executions: 50 held-out tasks across five folds,
repeated three times. Table~\ref{tab:collaboration-errors} compares the
initial MAS, Meta-Harness, and MASH. Percentages are
$100\,C_{m,\ell}/|D_m|$, dividing confirmed issue counts by all
non-perfect executions, including unknown labels.

\begin{table}[H]
\centering
\small
\setlength{\tabcolsep}{4pt}
\renewcommand{\arraystretch}{1.08}
\caption{
Confirmed issues in centralized Finance-Agent test executions:
count (percentage conditional on a non-perfect execution for that method).
Executions count once per level; the edge-or-graph union is deduplicated.
Unknown labels remain in the denominator; their counts appear in
Table~\ref{tab:failure-label-unknown}.
}
\label{tab:collaboration-errors}
\begin{tabular}{@{}lrrr@{}}
\toprule
Metric & \shortstack{MAS-\\Centralized}
& \shortstack{Meta-\\Harness} & MASH \\
\midrule
Non-perfect executions & 117 & 101 & 91 \\
\midrule
Node
& 56 (47.9\%) & 36 (35.6\%) & 32 (35.2\%) \\
Edge
& 42 (35.9\%) & 32 (31.7\%) & 20 (22.0\%) \\
Graph
& 45 (38.5\%) & 36 (35.6\%) & 21 (23.1\%) \\
Edge or graph (union)
& 66 (56.4\%) & 49 (48.5\%) & 31 (34.1\%) \\
\bottomrule
\end{tabular}
\end{table}

MASH reduces all three issue counts, most at edge and graph.
The edge-or-graph union falls from 49 to 31 executions after
deduplicating overlapping labels.

\appendixneedspace{0.32\textheight}
\subsubsection{Cross-Benchmark Comparison}

Table~\ref{tab:collaboration-errors-four-benchmarks} supplies the counts
behind Table~\ref{tab:main-level-errors}. For benchmark $b$ and level
$\ell$, the Meta-Harness and MASH issue counts define the relative
reduction
\[
r_{b,\ell}
=
\frac{
C^{\mathrm{Meta}}_{b,\ell}
-
C^{\mathrm{MASH}}_{b,\ell}
}{
C^{\mathrm{Meta}}_{b,\ell}
}
\times 100\%.
\]
For each level, we then report the unweighted macro-average
\[
\bar r_{\ell}
=
\frac{1}{4}\sum_b r_{b,\ell}.
\]

\begin{table}[H]
\centering
\small
\setlength{\tabcolsep}{4.0pt}
\renewcommand{\arraystretch}{1.08}
\caption{
Confirmed issue-count changes from Meta-Harness to MASH across four
benchmarks under the centralized workflow. Each cell reports
Meta-Harness $\rightarrow$ MASH followed by the relative reduction;
the final row is the unweighted macro-average across benchmarks.
Reductions use confirmed issue counts; Table~\ref{tab:collaboration-errors}
separately reports percentages within each method's non-perfect executions.
}
\label{tab:collaboration-errors-four-benchmarks}
\begin{tabular*}{\linewidth}{@{\extracolsep{\fill}}lccc@{}}
\toprule
Benchmark & Node & Edge & Graph \\
\midrule
Finance-Agent
& $36\!\rightarrow\!32$ (11.1\%)
& $32\!\rightarrow\!20$ (37.5\%)
& $36\!\rightarrow\!21$ (41.7\%) \\
BrowseComp-Plus
& $113\!\rightarrow\!108$ (4.4\%)
& $78\!\rightarrow\!56$ (28.2\%)
& $74\!\rightarrow\!55$ (25.7\%) \\
Plancraft
& $85\!\rightarrow\!77$ (9.4\%)
& $35\!\rightarrow\!32$ (8.6\%)
& $65\!\rightarrow\!50$ (23.1\%) \\
WorkBench
& $87\!\rightarrow\!82$ (5.7\%)
& $54\!\rightarrow\!39$ (27.8\%)
& $89\!\rightarrow\!66$ (25.8\%) \\
\midrule
Macro average
& \textbf{7.7\%}
& \textbf{25.5\%}
& \textbf{29.1\%} \\
\bottomrule
\end{tabular*}
\end{table}

MASH has lower confirmed issue counts at every level on all four benchmarks,
with larger macro-averaged reductions at edge and graph than at node.

\begingroup
\paragraph{Comparison with HarnessX.}
Table~\ref{tab:harnessx-failure-patterns} extends the frozen-harness
failure-pattern analysis to HarnessX across all four benchmarks under
the centralized workflow. Each percentage uses the corresponding
method's non-perfect execution count as its denominator.
Using the same unweighted macro-average of benchmark-level relative
count reductions defined above, MASH reduces node-, edge-, and
graph-level issue counts relative to HarnessX by 2.5\%, 17.0\%,
and 24.7\%, respectively.

The reductions vary across benchmarks and levels. MASH has slightly
more node-level issues on BrowseComp-Plus (108 versus 106) and
WorkBench (82 versus 81), and more edge-level issues on Plancraft
(32 versus 30), while its graph-level issue count is lower on all
four benchmarks. Counts and within-method percentages can also move
in different directions: on Finance-Agent, the node count decreases
from 34 to 32, but the percentage increases from 34.3\% to 35.2\%
because the non-perfect execution population decreases from 99 to 91.
\endgroup

\begin{table}[htbp]
\centering
\caption{Failure-pattern comparison of HarnessX and MASH under the
centralized workflow. Issue cells report confirmed counts and their
percentages among that method's non-perfect executions. Levels may
overlap. The last row reports the unweighted macro-average relative
reduction in issue counts from HarnessX to MASH across benchmarks.}
\label{tab:harnessx-failure-patterns}
\begingroup
\MashTableSetup
\begin{tabular*}{\linewidth}{@{\extracolsep{\fill}}lrrrr@{}}
\toprule
Method & \shortstack{Non-perfect\\executions} & Node & Edge & Graph \\
\midrule
\multicolumn{5}{l}{\textbf{Finance-Agent}} \\
HarnessX & 99 & 34 (34.3\%) & 29 (29.3\%) & 33 (33.3\%) \\
MASH & 91 & 32 (35.2\%) & 20 (22.0\%) & 21 (23.1\%) \\
\midrule
\multicolumn{5}{l}{\textbf{BrowseComp-Plus}} \\
HarnessX & 224 & 106 (47.3\%) & 73 (32.6\%) & 70 (31.2\%) \\
MASH & 213 & 108 (50.7\%) & 56 (26.3\%) & 55 (25.8\%) \\
\midrule
\multicolumn{5}{l}{\textbf{Plancraft}} \\
HarnessX & 132 & 83 (62.9\%) & 30 (22.7\%) & 63 (47.7\%) \\
MASH & 115 & 77 (67.0\%) & 32 (27.8\%) & 50 (43.5\%) \\
\midrule
\multicolumn{5}{l}{\textbf{WorkBench}} \\
HarnessX & 170 & 81 (47.6\%) & 49 (28.8\%) & 83 (48.8\%) \\
MASH & 156 & 82 (52.6\%) & 39 (25.0\%) & 66 (42.3\%) \\
\midrule
\multicolumn{2}{l}{Macro count reduction}
& 2.5\% & 17.0\% & 24.7\% \\
\bottomrule
\end{tabular*}
\endgroup
\end{table}

\clearpage
\section{Search Efficiency and Resource Accounting}
\label{app:optimization-efficiency}
We first compare filtering stages and complete search strategies,
then account for scheduled evaluations and actual development use.
The measurements have different units and aggregation scopes:
\begin{itemize}[leftmargin=*,itemsep=0.2em]
  \item \textbf{MAS task starts (task attempts):} launched system
  executions, including failures and whole-task retries. Development
  totals include initialization; the per-round strategy comparison
  excludes it. Both exclude final testing.
  \item \textbf{Total LLM calls:} requests from MAS executions and the
  optimizer, reported per round in the filtering ablation.
  \textbf{Optimizer requests} count only optimizer-side model calls
  in the baseline compute table.
  \item \textbf{Optimizer tokens:} request-level input and output
  usage, aggregated separately from MAS and benchmark-grader usage.
\end{itemize}
Candidate counts retain the definitions in
Appendix~\ref{app:candidate-quality-cost}: a submitted version,
a retained candidate, and an accepted revision are different events.

\subsection{Filtering-Stage Ablations}
\label{app:check-ablation-protocol}
\label{app:filtering-stage-contributions}

We compare Full MASH with three variants that each remove one filtering
stage, measuring final performance and total search cost.
Diagnosis, generation, editable scope, and adoption follow the common
protocol in Appendix~\ref{app:ablation-protocols}. The optimizer is GPT-5.4, with session caps of 11 for
analysis and 13 for generation. Only the indicated stage and its
associated feedback are removed; realized LLM request counts may differ.

\begin{itemize}
    \item \textbf{w/o Stage 1} removes the explicit rule-based
    syntax and semantics check and its associated interception,
    repair, and feedback.
    Full MASH's generation requirements remain unchanged; runtime failures
    are still recorded under the common execution-failure handling.
    \item \textbf{w/o Stage 2} removes the effectiveness check,
    including the fixed four-task executions, the $0.5$-point threshold
    decision, and the associated filtering feedback.
    Stage-1 survivors proceed directly to Stage 3.
    \item \textbf{w/o Stage 3} removes the robustness check,
    including the eight-task executions, the $0.25$-point threshold
    decision, and the associated filtering feedback.
    Stage-2 survivors proceed directly to paired evaluation in the
    next round.
\end{itemize}

Surviving candidates follow the remaining checks and the same
32-task, two-repetition adoption comparison. After search, each fold's
selected harness is frozen. For each of three test repetitions, we
average scores over the 50 held-out tasks pooled across five folds,
then report the mean of the three repetition-level scores.
Table~\ref{tab:filtering-ablation} reports these scores alongside
total LLM calls per optimization round.

\begin{table}[H]
\centering
\small
\renewcommand{\arraystretch}{1.08}
\caption{Filtering-stage ablations on Finance-Agent under the centralized
workflow. Scores are mean test scores ($\times100$). Total LLM calls per
round include model requests during MAS executions and optimizer requests
for diagnosis, generation, and repair, including failed attempts and retries. Call counts are averaged over
100 optimization rounds across five folds and exclude initialization and
final testing. Call change is relative to Full MASH.}
\label{tab:filtering-ablation}
\begin{tabular*}{0.96\linewidth}{@{\extracolsep{\fill}}lrrr@{}}
\toprule
Variant & Score & \shortstack{LLM calls\\per round} & Call change \\
\midrule
Full MASH & \textbf{52.88} & 13,447.3 & -- \\
w/o Stage 1 & 52.43 & 14,757.7 & $+9.7\%$ \\
w/o Stage 2 & 52.67 & 15,102.7 & $+12.3\%$ \\
w/o Stage 3 & 51.96 & 13,223.8 & $-1.7\%$ \\
\bottomrule
\end{tabular*}
\end{table}

In these ablations, Stages 1 and 2 improve final performance while
reducing total search calls: removing them increases calls by 9.7\%
and 12.3\%, respectively, while lowering scores. Removing the
robustness check (Stage 3) saves 1.7\% of calls but produces the
largest score drop. Appendix~\ref{app:candidate-evaluation-cost}
separates planned task starts avoided by filtering from realized
development task starts.

\clearpage
\subsection{Search Progress, Resource Use, and Sensitivity}
\label{app:strategy-resources}
\label{app:diagnosis-protocol}

\subsubsection{Search Progress and Resource Use}

This analysis compares search trajectories for F/U/O/R/L and per-round
resources for F/R/L on centralized Finance-Agent. All start from $H_0$
and use 20 rounds under Appendix~\ref{app:ablation-protocols}; the
earliest parent update is in round 2.

\paragraph{Search trajectories: F/U/O/R/L.}
Each strategy is optimized independently within each fold, with an initial
candidate cap of three and 4,000 development task attempts. Adoption uses
development scores. End-of-round harnesses are saved and evaluated on the
corresponding held-out test fold after search. These post-hoc scores are
used only for plotting, not for candidate selection, adoption, or early stopping.

\begin{figure}[H]
\centering
\includegraphics[width=\linewidth]{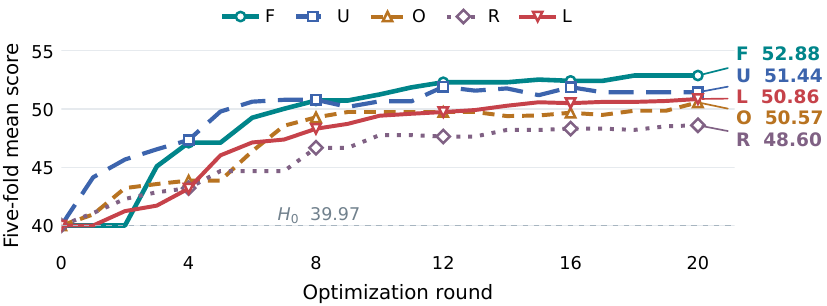}
\caption{
Search trajectories on centralized Finance-Agent. Scores are five-fold
means of three test repetitions per checkpoint ($\times 100$), evaluated post hoc;
the horizontal line marks $H_0=39.97$.
F (MASH): structured diagnosis with cross-level candidates;
U: structured, single-level; O: free-form, cross-level;
R: free-form, single-level; L: level-wise search.
Final F/U/O/R scores correspond to Run 1 in Table~\ref{tab:component-ablation-runs}.
The x-axis indexes progress by logged optimization round; realized
resource use is reported separately in Table~\ref{tab:failure-diagnosis}.
}
\label{fig:organization-search-curves}
\end{figure}

U improves faster early in search, whereas F catches up and finishes
highest (Figure~\ref{fig:organization-search-curves}). L finishes below
U but above O and R. Flat segments indicate unchanged five-fold means
at logged checkpoints; adoption requires a positive gain in the
next-round paired evaluation.

\paragraph{Per-round resources: F/R/L.}
F/R/L compare complete strategies with different evidence analysis,
revision organization, and implementation requirements
(Appendix~\ref{app:organization-protocol}). They share the round schedule,
analysis and generation allowances, and access to raw evidence. The parent
$\bar H_t$ stays fixed during each pool evaluation. Realized resource use differs:
Table~\ref{tab:failure-diagnosis} separates MAS starts from optimizer
tokens for analysis, generation and repair, evaluation, and selection.

\begin{table}[H]
\centering
\small
\setlength{\tabcolsep}{4pt}
\renewcommand{\arraystretch}{1.08}
\caption{
Performance and per-round resources for F (MASH), R (free-form,
single-level search), and L (level-wise search) on centralized
Finance-Agent. Test scores ($\times 100$) average three repetitions.
Resources average all 100 rounds across five folds, including those
without an update; initialization and final testing are excluded.
}
\label{tab:failure-diagnosis}
\begin{tabularx}{0.96\linewidth}{@{}p{0.48\linewidth}*{3}{>{\raggedleft\arraybackslash}X}@{}}
\toprule
Metric & F (MASH) & R & L \\
\midrule
Test score & \textbf{52.88} & 48.60 & 50.86 \\
MAS task starts per round & 166.3 & 145.4 & 159.7 \\
Optimizer tokens per round (k) & 90.3 & 88.9 & 99.9 \\
\bottomrule
\end{tabularx}
\end{table}

\begin{samepage}
MASH has the highest score. Relative to L, it uses fewer optimizer
tokens but more MAS starts per round (166.3 versus 159.7). R uses the
fewest resources on both measures and scores lowest.\par
\end{samepage}

\begingroup
\paragraph{Cumulative resources to reach target scores.}
Table~\ref{tab:target-score-resources} compares MASH, Meta-Harness,
and HarnessX on Finance-Agent at target scores of 40, 44, and 47
on the 0--100 scale. For each target, we report cumulative MAS
task attempts and optimizer tokens, summed across the five folds.
These are cumulative resource totals, whereas
Table~\ref{tab:failure-diagnosis} reports average costs per
optimization round for the diagnosis-strategy variants.

\paragraph{Checkpoint evaluation and cost accounting.}
The targets refer to the mean held-out test score across the five
folds. At each saved checkpoint $k$, we freeze that checkpoint's
harness separately in each fold and evaluate it on the held-out
test tasks three times. Let $S^{\mathrm{test}}_{f,k}$ denote the
mean test score over these repetitions in fold $f$.
For each target $\tau\in\{40,44,47\}$, we identify the first saved
checkpoint whose five-fold mean reaches the target:
\[
k^*(\tau)=\min\left\{k:
\frac{1}{5}\sum_{f=1}^{5}S^{\mathrm{test}}_{f,k}\geq\tau
\right\}.
\]
These test evaluations are performed retrospectively on saved
checkpoints and do not participate in harness selection.
For either resource measure, let $C_{f,k}$ be its cumulative
development cost in fold $f$ through checkpoint $k$. We report
\[
C(\tau)=\sum_{f=1}^{5}C_{f,k^*(\tau)},
\]
using the same checkpoint index for all five folds and computing
the total separately for MAS task attempts and optimizer tokens.
Development MAS attempts exclude the retrospective test executions.
Optimizer tokens sum only the input and output tokens of optimizer
requests; executor and grader tokens are excluded from this total.

MASH uses fewer optimizer tokens than both baselines at all three
targets. It also uses fewer MAS task attempts at targets 44 and 47.
At target 47, MASH uses 3,480 attempts and 1.73 million optimizer
tokens, compared with HarnessX's 13,218 attempts and 38.72 million
tokens, reductions of 73.7\% and 95.5\%, respectively.
At target 40, HarnessX uses fewer attempts than MASH
(1,865 versus 2,210), while MASH uses fewer optimizer tokens
(1.13 versus 4.96 million).
\endgroup

\begin{table}[htbp]
\centering
\caption{Cumulative resources required to reach each target score
on Finance-Agent, summed over five folds. Target scores are on the
0--100 scale; optimizer tokens are in millions. Bold denotes the
lowest resource use in each column.}
\label{tab:target-score-resources}
\begingroup
\MashTableSetup
\begin{tabular*}{\linewidth}{@{\extracolsep{\fill}}lrrrrrr@{}}
\toprule
& \multicolumn{3}{c}{MAS task attempts}
& \multicolumn{3}{c}{Optimizer tokens (M)} \\
\cmidrule(lr){2-4}\cmidrule(lr){5-7}
Method & $\geq 40$ & $\geq 44$ & $\geq 47$
& $\geq 40$ & $\geq 44$ & $\geq 47$ \\
\midrule
Meta-Harness & 2,318 & 7,684 & 16,294 & 7.96 & 25.73 & 59.31 \\
HarnessX & \textbf{1,865} & 7,026 & 13,218 & 4.96 & 19.94 & 38.72 \\
MASH & 2,210 & \textbf{2,210} & \textbf{3,480}
& \textbf{1.13} & \textbf{1.13} & \textbf{1.73} \\
\bottomrule
\end{tabular*}
\endgroup
\end{table}

\clearpage
\subsubsection{Hyperparameter Sensitivity}
\label{app:hyperparameter-sensitivity}
We vary the round limit and initial candidate cap for full MASH on
centralized Finance-Agent. For the round sweep, each fold completes one
30-round search with an extended development budget and a candidate cap of
three. Saved checkpoints are tested after search. For the cap sweep,
each cap is run separately with a 20-round limit and the default 4,000
task attempts per fold, stopping if the remaining budget cannot support
continuation. Initial candidates and revisions share a call allowance;
the cap bounds initial proposals, not their realized count. Both sweeps
report five-fold means of three test repetitions per evaluated harness
after search, without using test feedback for optimization.

\begin{figure}[H]
    \centering
    \includegraphics[width=\linewidth]{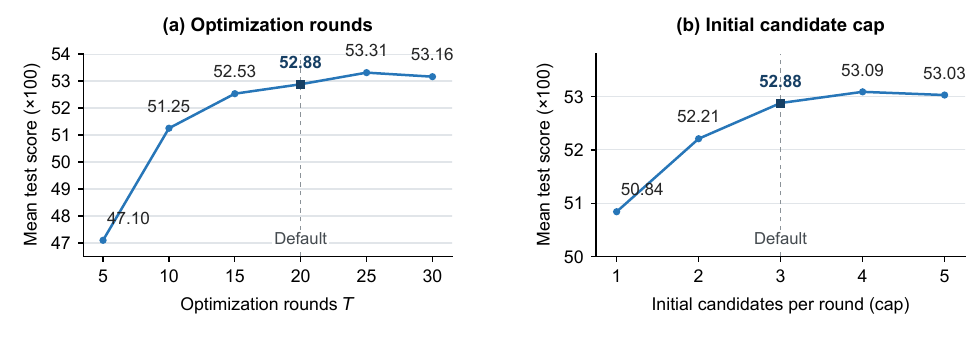}
    \caption{
        Hyperparameter sensitivity of full MASH on Finance-Agent
        under the centralized workflow using five-fold evaluation.
        (a) Checkpoints from one 30-round search per fold with an extended
        development budget and initial candidate cap of three.
        (b) Initial candidate caps of $\{1,2,3,4,5\}$
        with a 20-round limit and 4,000 development task attempts per fold.
        Scores are multiplied by 100. The marked defaults---20 optimization
        rounds and an initial candidate cap of three---capture most observed gains.
    }
    \label{fig:hyperparameter-sensitivity}
\end{figure}

\begingroup
\paragraph{Filtering thresholds.}
We examine the effectiveness threshold
$\theta_{\mathrm{eff}}$ and robustness threshold $\theta_{\mathrm{gen}}$
on all four benchmarks. For each benchmark, we vary one threshold
while fixing the other at its default value and keeping the screening-subset
size at the value listed in Table~\ref{tab:filtering-hyperparameters}.
Table~\ref{tab:filtering-threshold-sensitivity} reports the resulting
test scores. The default-setting scores correspond to Optimization Run 1,
whereas Table~\ref{tab:main-results} summarizes
three independent optimization runs at the defaults.
Relative to the selected defaults, the largest score decreases among
the tested perturbations are 1.16, 1.33, 1.34, and 1.67 points on
Finance-Agent, BrowseComp-Plus, Plancraft, and WorkBench, respectively.
The threshold-selection procedure is described in
Appendix~\ref{app:data-splits}.
\endgroup

\begin{table}[t]
\centering
\caption{Sensitivity to the effectiveness and robustness thresholds.
Test scores are multiplied by 100. Each non-default row varies one
threshold with the other fixed; the selected default configuration
for each benchmark is bold.}
\label{tab:filtering-threshold-sensitivity}
\begingroup
\MashTableSetup
\begin{tabular*}{\linewidth}{@{\extracolsep{\fill}}lccc@{}}
\toprule
Benchmark & $\theta_{\mathrm{eff}}$ & $\theta_{\mathrm{gen}}$
& Test score \\
\midrule
Finance-Agent & 0.000 & 0.0025 & 52.16 \\
& \textbf{0.005} & \textbf{0.0025} & \textbf{52.88} \\
& 0.010 & 0.0025 & 51.72 \\
& 0.005 & 0.0000 & 52.29 \\
& 0.005 & 0.0050 & 51.83 \\
\midrule
BrowseComp-Plus & 0.000 & 0.075 & 28.33 \\
& \textbf{0.060} & \textbf{0.075} & \textbf{29.00} \\
& 0.190 & 0.075 & 27.67 \\
& 0.060 & 0.025 & 28.67 \\
& 0.060 & 0.125 & 28.00 \\
\midrule
Plancraft & 0.040 & 0.025 & 61.00 \\
& \textbf{0.100} & \textbf{0.025} & \textbf{61.67} \\
& 0.190 & 0.025 & 60.33 \\
& 0.100 & 0.000 & 61.33 \\
& 0.100 & 0.075 & 60.67 \\
\midrule
WorkBench & 0.050 & 0.075 & 47.67 \\
& \textbf{0.150} & \textbf{0.075} & \textbf{48.00} \\
& 0.250 & 0.075 & 46.33 \\
& 0.150 & 0.025 & 47.33 \\
& 0.150 & 0.125 & 47.00 \\
\bottomrule
\end{tabular*}
\endgroup
\end{table}

\clearpage
\subsection{Candidate Evaluation Costs and Task-Start Accounting}
\label{app:candidate-evaluation-cost}

\subsubsection{Scheduled Evaluations and Cascade Savings}
A complete evaluation path for one F/U/O/R/L version allocates
152 MAS task starts: $8=2\times4$ for Stage 2,
$16=2\times8$ for Stage 3, and $128=2\times32\times2$ for
next-round adoption. The factors count both harnesses, tasks, and
repetitions. Stage 1 uses no MAS executions; rejection avoids later stages.

Let $N$ count submitted versions and $S$ scheduled starts avoided by
filtering-stage rejection. Table~\ref{tab:constraint-evaluation-cost}
reports the avoided fraction $S/(152N)$ and actual development starts,
pooled across five folds.

\begin{table}[H]
\centering
\small
\setlength{\tabcolsep}{6pt}
\renewcommand{\arraystretch}{1.08}
\caption{
Candidate costs across five centralized Finance-Agent folds.
$N$: submitted versions; $S$: scheduled starts avoided by filtering;
avoided fraction: $S/(152N)$. Development counts actual launches,
excluding final testing. Variants: Table~\ref{tab:organization-variants}.
}
\label{tab:constraint-evaluation-cost}
\begin{tabular}{lrrrr}
\toprule
Variant & $N$
& \shortstack{Development\\task starts}
& \shortstack{Avoided\\task starts $S$}
& \shortstack{Avoided\\(\%)} \\
\midrule
F (MASH) & 164 & 16,946 &  9,640 & 38.7 \\
U        & 173 & 15,820 & 11,160 & 42.4 \\
O        & 169 & 15,191 & 11,768 & 45.8 \\
R        & 181 & 14,868 & 13,704 & 49.8 \\
L        & 176 & 16,292 & 11,600 & 43.4 \\
\bottomrule
\end{tabular}
\end{table}

Filtering avoids 38.7--49.8\% of the scheduled allocation; the ledger
below accounts for unfinished comparisons and other launches.

\appendixneedspace{0.24\textheight}
\subsubsection{Actual Task Starts}
\label{app:cost-reconciliation}
A slot schedules one harness on one task and repetition. First, subtract
filtering savings $S$ from $152N$. Next, subtract unfinished next-round slots
($D_{\mathrm{inc}}$) and add back their one-sided launches ($C_{\mathrm{unp}}$);
$Q-M$ counts unfinished comparisons. Finally, add first parent-only diagnosis
launches when no candidate proceeds ($C_{\mathrm{parent}}$), initialization
and its retries ($C_{\mathrm{init}}$), and other whole-task retries,
including parent-only retries ($C_{\mathrm{retry}}$):

\begin{equation}
C_{\mathrm{dev}}
=152N-S-D_{\mathrm{inc}}+C_{\mathrm{unp}}
 +C_{\mathrm{parent}}+C_{\mathrm{init}}+C_{\mathrm{retry}}.
\label{eq:development-start-reconciliation}
\end{equation}
Table~\ref{tab:cost-reconciliation} reconciles the totals for all five
strategies; final testing is excluded.

\begin{table}[H]
\centering
\small
\setlength{\tabcolsep}{5pt}
\renewcommand{\arraystretch}{1.08}
\caption{
Development task-start ledger across five Finance-Agent folds.
Signed adjustments follow Equation~\ref{eq:development-start-reconciliation}; residual
is the reconciled total minus Table~\ref{tab:constraint-evaluation-cost}.
}
\label{tab:cost-reconciliation}
\begin{tabularx}{\linewidth}{@{}p{0.39\linewidth}*{5}{>{\raggedleft\arraybackslash}X}@{}}
\toprule
Accounting item & F & U & O & R & L \\
\midrule
Complete schedule $152N$
& 24,928 & 26,296 & 25,688 & 27,512 & 26,752 \\
Filtering adjustment $-S$
& $-9{,}640$ & $-11{,}160$ & $-11{,}768$ & $-13{,}704$ & $-11{,}600$ \\
Remaining allocation $152N-S$
& 15,288 & 15,136 & 13,920 & 13,808 & 15,152 \\
\addlinespace[3pt]
Incomplete slots $-D_{\mathrm{inc}}$
& $-432$ & $-648$ & $-556$ & $-512$ & $-382$ \\
Unpaired starts $(+C_{\mathrm{unp}})$
& 9 & 10 & 8 & 10 & 10 \\
Parent-only starts $(+C_{\mathrm{parent}})$
& 1,664 & 896 & 1,408 & 1,152 & 1,088 \\
Initialization $(+C_{\mathrm{init}})$
& 320 & 322 & 322 & 328 & 320 \\
Whole-task retries $(+C_{\mathrm{retry}})$
& 97 & 104 & 89 & 82 & 104 \\
\midrule
Development starts $C_{\mathrm{dev}}$
& 16,946 & 15,820 & 15,191 & 14,868 & 16,292 \\
Residual & 0 & 0 & 0 & 0 & 0 \\
\bottomrule
\end{tabularx}
\end{table}

The net unstarted counts $D_{\mathrm{inc}}-C_{\mathrm{unp}}$ are 423,
638, 548, 502, and 372 for F/U/O/R/L. Per-round summaries in
Appendix~\ref{app:strategy-resources} additionally exclude initialization.

\appendixneedspace{0.35\textheight}
\subsection{Development Compute Across Baselines}
\label{app:baseline-compute}
Table~\ref{tab:baseline-development-compute} compares development use
under the task-attempt budgets in Appendix~\ref{app:baseline-details}.
The cap bounds launched MAS executions; optimizer requests and tokens
measure the separate cost of proposing and assessing changes.

\begin{table}[H]
\centering
\small
\renewcommand{\arraystretch}{1.08}
\setlength{\tabcolsep}{5pt}
\caption{Development compute on Finance-Agent under the centralized
workflow, aggregated over five folds. Each optimization method has an
independent cap of 4,000 MAS task attempts per fold, with no transfer
between folds. Input and output are optimizer tokens in millions;
final testing is excluded.}
\label{tab:baseline-development-compute}
\begin{tabular}{lrrrr}
\toprule
Method & Task attempts & Optimizer calls & Input (M) & Output (M) \\
\midrule
MAS-Centralized & 0      & 0      & 0.00   & 0.00  \\
GEPA            & 19,742 & 2,147  & 26.31  & 3.41  \\
RHI             & 3,487  & 3,287  & 18.94  & 1.34  \\
ACE             & 3,118  & 2,241  & 41.76  & 2.07  \\
AHE             & 3,276  & 12,083 & 151.37 & 12.64 \\
Life-Harness    & 6,914  & 873    & 17.68  & 1.37  \\
Meta-Harness    & 17,284 & 2,317  & 58.42  & 4.71  \\
HarnessX        & 17,533 & 2,286  & 48.11  & 4.18  \\
Meta-Team       & 236    & 2,296  & 38.87  & 1.84  \\
AFlow           & 13,684 & 131    & 2.71   & 0.34  \\
MASH            & 16,946 & 913    & 8.27   & 0.93  \\
\bottomrule
\end{tabular}
\end{table}

The methods allocate computation differently: AHE uses relatively few
MAS executions but many optimizer requests, whereas AFlow evaluates
many executions per request. Meta-Team makes approximately one
development pass per fold, plus retries and auxiliary evaluations;
its per-task evolution produces more optimizer requests than task
attempts. GEPA approaches the cap, while other methods may stop earlier
under their native schedules.

MAS-Centralized has zero development optimization cost because it
performs no search. Its frozen test executions are excluded under the
same accounting convention.

\begingroup
\paragraph{Comparisons under extended search budgets.}
To examine the effect of the differences in realized MAS execution
counts, we extend development search for RHI, ACE, AHE, Life-Harness,
and AFlow on Finance-Agent under the centralized workflow.
Table~\ref{tab:extended-execution-comparison} reports their original
and extended task-attempt counts and test scores, alongside the
unchanged reference results for the other methods.
Task attempts are aggregated over the five folds using the accounting
convention above, excluding final testing.
These comparisons use the Optimization Run 1 results as their
reference; Table~\ref{tab:main-results} summarizes
three independent optimization runs.

For these extended searches, the cap of 4,000 MAS task attempts
remains enforced independently in each fold. Each method retains
its original rule for selecting the final harness.

We extend RHI by increasing the number of harness revisions and
ACE by increasing adaptation epochs, with the number of per-sample
reflection iterations unchanged. For AHE, we increase evolution
iterations while retaining the historical best version and rollback
mechanism. For Life-Harness, we increase the number of revisions
while retaining its original improvement-adoption rule. For AFlow,
we increase the number of Monte Carlo tree search expansions.

The extended methods use 16,854--17,038 task attempts, close to
MASH's 16,946. RHI, AHE, Life-Harness, and AFlow improve by
1.11, 0.75, 1.26, and 0.61 score points, respectively, whereas
ACE decreases by 0.34 points.
MASH attains 52.88, exceeding HarnessX's 48.76 by 4.12 points
at similar execution counts. GEPA is retained as a reference at its
higher original count of 19,742 attempts.
Thus, extending the baselines with fewer development task attempts does not close the
observed performance gap in this comparison.
\endgroup

\begin{table}[t]
\centering
\caption{Finance-Agent results before and after extending development
search for the baselines with fewer development task attempts. Task attempts are aggregated
over five folds; test scores are multiplied by 100.
The search budgets for the other methods remain unchanged.}
\label{tab:extended-execution-comparison}
\begingroup
\MashTableSetup
\begin{tabular*}{\linewidth}{@{\extracolsep{\fill}}lrrrrr@{}}
\toprule
& \multicolumn{2}{c}{Task attempts} & \multicolumn{2}{c}{Test score}
& $\Delta$ Score \\
\cmidrule(lr){2-3}\cmidrule(lr){4-5}
Method & Original & Extended & Original & Extended & \\
\midrule
MAS-Centralized & 0 & 0 & 39.97 & 39.97 & 0.00 \\
GEPA & 19,742 & 19,742 & 39.32 & 39.32 & 0.00 \\
RHI & 3,487 & 16,911 & 43.02 & 44.13 & $+1.11$ \\
ACE & 3,118 & 17,038 & 38.23 & 37.89 & $-0.34$ \\
AHE & 3,276 & 16,854 & 46.83 & 47.58 & $+0.75$ \\
Life-Harness & 6,914 & 16,987 & 43.49 & 44.75 & $+1.26$ \\
Meta-Harness & 17,284 & 17,284 & 47.24 & 47.24 & 0.00 \\
HarnessX & 17,533 & 17,533 & 48.76 & 48.76 & 0.00 \\
AFlow & 13,684 & 16,863 & 42.85 & 43.46 & $+0.61$ \\
\textbf{MASH} & 16,946 & 16,946 & \textbf{52.88}
& \textbf{52.88} & 0.00 \\
\bottomrule
\end{tabular*}
\endgroup
\end{table}

\begingroup
\paragraph{Comparison at 236 task attempts.}
Meta-Team's native schedule makes approximately one development pass
over the 32 diagnosis tasks in each fold, with an evolution step after
each task. We therefore compare it with MASH separately at the same
total of 236 task attempts, rather than extending Meta-Team to the
execution counts in Table~\ref{tab:extended-execution-comparison}.
Table~\ref{tab:low-execution-comparison} shows that MASH scores 42.49
versus Meta-Team's 41.86, a gain of 0.63 points at this execution count.
\endgroup

\begin{table}[t]
\centering
\caption{Finance-Agent comparison at the same total of 236
development task attempts across five folds. Final testing is excluded
from the attempt count; test scores are multiplied by 100.}
\label{tab:low-execution-comparison}
\begingroup
\MashTableSetup
\begin{tabular*}{\linewidth}{@{\extracolsep{\fill}}lrr@{}}
\toprule
Method & Task attempts & Test score \\
\midrule
Meta-Team & 236 & 41.86 \\
\textbf{MASH} & 236 & \textbf{42.49} \\
\bottomrule
\end{tabular*}
\endgroup
\end{table}

\clearpage
\section{Generalization}
\label{app:generalization-protocols}
We evaluate harness optimization under fixed workflows and transfer to
new execution backbones.

\subsection{Workflow Generalization}
\label{app:workflow-protocol}
\label{app:workflow-results}

We optimize independently under centralized, decentralized, hybrid,
and independent workflows, editing each workflow's native decision
points while keeping its topology and communication channels fixed. The independent workflow permits local execution and terminal
aggregation edits, without intermediate communication or shared
intermediate agent state.

Table~\ref{tab:workflow-generalization} compares MASH with the initial
MAS and strongest reported baseline in each benchmark--workflow setting.

\begin{table}[H]
\centering
\small
\setlength{\tabcolsep}{4.0pt}
\renewcommand{\arraystretch}{1.08}
\caption{
Harness optimization under four fixed workflows, with a separate search
for each workflow (scores
$\times 100$; mean over three frozen-harness test repetitions).
Best baseline is the highest score among the nine optimization
baselines from the main comparison, each adapted to the
corresponding fixed workflow, for each benchmark--workflow pair.
HX, MH, and AF denote HarnessX, Meta-Harness,
and AFlow. Slashes denote ties, bold marks the best score, and Improvement
is MASH minus the best baseline in score points. Centralized results are
from Optimization Run 1; Table~\ref{tab:main-results}
reports means over three independent optimization runs.
}
\label{tab:workflow-generalization}
\begin{tabular*}{\linewidth}{@{\extracolsep{\fill}}lcccc@{}}
\toprule
Method
& Finance-Agent & BrowseComp-Plus & Plancraft & WorkBench \\
\midrule

\multicolumn{5}{l}{\textbf{Centralized}} \\
MAS-Centralized
& 39.97 & 20.00 & 53.00 & 35.00 \\
Best baseline
& 48.76 (HX)
& 25.33 (HX)
& 57.00 (AF)
& 43.33 (HX) \\
\textbf{MASH}
& \textbf{52.88}
& \textbf{29.00}
& \textbf{61.67}
& \textbf{48.00} \\
\quad Improvement
& +4.12 & +3.67 & +4.67 & +4.67 \\
\midrule

\multicolumn{5}{l}{\textbf{Decentralized}} \\
MAS-Decentralized
& 38.26 & 23.67 & 53.00 & 41.00 \\
Best baseline
& 46.25 (HX)
& 29.33 (MH/AF)
& 56.67 (AF)
& 49.33 (HX) \\
\textbf{MASH}
& \textbf{48.97}
& \textbf{34.67}
& \textbf{60.33}
& \textbf{53.33} \\
\quad Improvement
& +2.72 & +5.34 & +3.66 & +4.00 \\
\midrule

\multicolumn{5}{l}{\textbf{Hybrid}} \\
MAS-Hybrid
& 39.55 & 21.67 & 55.00 & 37.00 \\
Best baseline
& 47.86 (HX)
& 28.00 (HX)
& 59.67 (AF)
& 44.33 (MH) \\
\textbf{MASH}
& \textbf{50.76}
& \textbf{32.33}
& \textbf{63.67}
& \textbf{49.67} \\
\quad Improvement
& +2.90 & +4.33 & +4.00 & +5.34 \\
\midrule

\multicolumn{5}{l}{\textbf{Independent}} \\
MAS-Independent
& 35.75 & 14.67 & 45.00 & 46.00 \\
Best baseline
& 40.71 (HX)
& 18.67 (HX)
& 47.33 (HX/AF)
& 48.67 (HX) \\
\textbf{MASH}
& \textbf{42.43}
& \textbf{24.00}
& \textbf{52.67}
& \textbf{52.00} \\
\quad Improvement
& +1.72 & +5.33 & +5.34 & +3.33 \\
\bottomrule
\end{tabular*}
\end{table}

MASH exceeds the best reported baseline in all 16 benchmark--workflow
pairs, by 1.72--5.34 points. No single workflow yields the highest MASH
score on every benchmark: centralized leads on Finance-Agent,
decentralized on BrowseComp-Plus and WorkBench, and hybrid on Plancraft.
The gains extend to the independent workflow without intermediate
communication.

\clearpage
\subsection{Backbone Transfer and Adaptation}
\label{app:backbone-protocol}

On centralized Finance-Agent, each fold transfers its development-selected
harness from Gemini 2.5 Flash-Lite to GPT-4o mini and Claude Haiku 4.5.
Table~\ref{tab:backbone-transfer-results} compares the initial harness
$H_0$, the unchanged source harness (\emph{Frozen transfer}), and
subsequent target-side optimization (\emph{Re-optimized}).

Roles, topology, tool permissions, scorer, per-task budgets, diagnosis
and generation models, and search rules remain fixed. Re-optimization
uses 20 additional rounds per target backbone, with at most three
initial candidates per round and session caps of 11 for analysis and
13 for generation. Adaptation uses only the corresponding fold's
diagnosis and validation partitions.

\begin{table}[H]
\centering
\small
\setlength{\tabcolsep}{7pt}
\renewcommand{\arraystretch}{1.08}
\caption{Backbone transfer from Gemini 2.5 Flash-Lite on centralized
Finance-Agent. Initial MAS uses $H_0$; frozen transfer applies the source
harness unchanged; re-optimization adds 20 target-side rounds. Scores
are means over three frozen-harness test repetitions ($\times 100$);
the best score per target is bold.}
\label{tab:backbone-transfer-results}
\begin{tabular}{lcc}
\toprule
Condition & GPT-4o mini & Claude Haiku 4.5 \\
\midrule
Initial MAS & 42.13 & 46.27 \\
Frozen transfer & 46.08 & 51.64 \\
Re-optimized & \textbf{53.36} & \textbf{54.11} \\
\bottomrule
\end{tabular}
\end{table}

Frozen transfer gains 3.95 points on GPT-4o mini and 5.37 on Claude
Haiku 4.5 over their initial MAS scores. Target-side re-optimization
adds another 7.28 and 2.47 points, respectively. Both backbones benefit
from transfer and further adaptation.

\begingroup
\paragraph{Control for transferred initialization.}
To evaluate the effect of initialization, we optimize $H_0$ and
the source-optimized harness $H_{\mathrm{source}}$ for 20 rounds
on each target backbone.
Table~\ref{tab:backbone-initialization-control} compares these two
conditions with the initial MAS and frozen transfer.
Source and target round counts are reported separately:
the transferred initialization includes 20 source-side rounds before
the 20 target-side rounds, whereas optimization from $H_0$ uses
only the 20 target-side rounds.

Without target-side optimization, frozen transfer improves the score
over the initial MAS by 3.95 points on GPT-4o mini and 5.37 points
on Claude Haiku 4.5. After 20 target-side rounds, optimization from
$H_0$ reaches 52.17 and 53.43, compared with 53.36 and 54.11 from
$H_{\mathrm{source}}$. The additional gains from transferred
initialization are therefore 1.19 and 0.68 points.
Thus, transferred initialization yields higher scores than optimizing
from $H_0$ after the same 20 target-side rounds, although this additional
gain is smaller than the gain from frozen transfer itself.
\endgroup

\begin{table}[htbp]
\centering
\caption{Cross-backbone initialization comparison on Finance-Agent
under the centralized workflow. Source rounds use Gemini 2.5 Flash-Lite;
target rounds use the model named in the score column.
Test scores are multiplied by 100. The final row compares the two
conditions with 20 target-side optimization rounds.}
\label{tab:backbone-initialization-control}
\begingroup
\MashTableSetup
\begin{tabular*}{\linewidth}{@{\extracolsep{\fill}}lrrrr@{}}
\toprule
Condition & \shortstack{Source\\rounds} & \shortstack{Target\\rounds}
& \shortstack{GPT-4o\\mini} & \shortstack{Claude\\Haiku 4.5} \\
\midrule
Initial MAS ($H_0$) & 0 & 0 & 42.13 & 46.27 \\
Frozen transfer ($H_{\mathrm{source}}$) & 20 & 0 & 46.08 & 51.64 \\
Optimize from $H_0$ & 0 & 20 & 52.17 & 53.43 \\
Optimize from $H_{\mathrm{source}}$ & 20 & 20 & 53.36 & 54.11 \\
\midrule
Gain from transferred initialization & --- & --- & $+1.19$ & $+0.68$ \\
\bottomrule
\end{tabular*}
\endgroup
\end{table}

\clearpage
\section{Failure Annotation and Measurement}
\label{app:failure-labeling-protocol}

This protocol defines the population, level-specific labels, review,
and aggregation for Appendix~\ref{app:collaboration-errors}. It counts
the occurrence of observable, level-specific process defects in
frozen-harness test executions.

\subsection{Population and Annotation Evidence}
\paragraph{Scope and unit of analysis.}
The unit is one test execution with a score below the maximum
(a \emph{non-perfect execution}). Before selecting non-perfect executions,
Finance-Agent supplies 150 test executions per method: 50 held-out tasks across five folds,
each repeated three times. The other benchmarks also use three
repetitions per held-out task (Appendix~\ref{app:experimental-setup}).

Each record identifies the method, benchmark, task, repetition, and,
for Finance-Agent, fold. Annotators judge the trajectories independently
of search-time diagnoses and other error taxonomies.

\paragraph{Evidence available to annotators.}
Annotators receive the task requirements and an ordered record of roles,
assigned requests, agent-visible model inputs, tool observations,
delivered contributions, and team decisions. Annotation inputs exclude
method identities, scores, reference answers, rubric judgments, and
other annotators' labels. Scores determine eligibility only.

Judgments cite events in context. Omission judgments use the
recipient's full input snapshot, including visible messages and
context blocks.

\subsection{Level Definitions and Label States}
Labels follow the responsibility of the defective decision, independent
of the acting agent, implementation file, or possible repair hook.
Table~\ref{tab:failure-level-definitions} defines the three levels
and their boundaries.

\begin{table}[H]
\centering
\small
\setlength{\tabcolsep}{5pt}
\renewcommand{\arraystretch}{1.08}
\caption{Operational failure definitions and boundary cases.
Labels follow the responsibility of the defective decision.}
\label{tab:failure-level-definitions}
\begin{tabularx}{\linewidth}{@{}>{\raggedright\arraybackslash}p{0.17\linewidth}>{\raggedright\arraybackslash}X@{}}
\toprule
Level & Definition, examples, and boundary \\
\midrule
Node: local execution &
Retrieval, extraction, reasoning, calculation, tool use, or verification
fails to satisfy the agent's assigned work given its inputs. Examples
are using an incorrect unit despite receiving the correct one,
miscalculating available values, and selecting incorrect tool arguments.
An unsuccessful search or tool-service failure alone does not establish
a node defect. \\
\addlinespace
Edge: requests and delivery &
A request or contribution fails to preserve information required for
the interaction. Examples include omitting an essential request
constraint, an available reporting period, or a source binding. A
delivery omission concerns information the sender had acquired.
Leaving necessary work unassigned is a graph defect rather than a
request-expression defect. \\
\addlinespace
Graph: team decisions &
Allocation, acceptance, aggregation, conflict resolution, coverage
checking, or stopping fails to satisfy the overall task. Examples are
leaving necessary work unassigned, failing to integrate an available
contribution, and stopping with required work unresolved. A prescribed
budget stop alone does not establish defective stopping. \\
\bottomrule
\end{tabularx}
\end{table}

\paragraph{Label states and overlap.}
Each level receives one state: \emph{present} for a defect supported
by locatable evidence; \emph{absent} for an inspectable process with
no such defect; or \emph{unknown} for insufficient or ambiguous
evidence, such as missing inputs or incomplete logs. Labels describe
level-specific processes, not answer correctness. Infrastructure
errors are recorded separately.

Multiple levels require independent evidence. Passing an upstream
calculation error unchanged does not itself establish an edge defect,
nor does an incorrect aggregate answer establish a graph defect.

\appendixneedspace{0.15\textheight}
\subsection{Annotation and Review}
Both annotators apply the same rules separately at each level. Their
outputs contain the anonymous execution identifier, each level's label,
a responsibility-specific reason, and evidence references.

\begin{enumerate}[leftmargin=*,label=\arabic*.,itemsep=0.4em]
  \item \textbf{Primary annotation.}
  GPT-5.5 labels the eligible non-perfect executions using
  the trajectory evidence specified above.

  \item \textbf{Independent review.}
  Task identifiers are sampled before inspecting primary labels:
  two held-out tasks per Finance-Agent fold (10 total), and 20 tasks
  independently for each other benchmark. Within a benchmark, the same
  sampled tasks define the review frame for MAS-Centralized, Meta-Harness,
  and MASH across all three repetitions. Within this frame,
  Gemini 2.5 Flash independently labels eligible non-perfect
  executions without access to the primary output.

  \item \textbf{Adjudication\enspace and\enspace boundary\enspace checks.}
  Disagreements undergo human adjudication against the original
  trajectory evidence.
  Labels unresolved at adjudication retain the \emph{unknown} state.
  Ten additional executions are randomly sampled from records with complete
  three-level agreement for evidence and boundary checks.

  Final failure-count analyses use the final labels after review.
\end{enumerate}

\begin{table}[H]
\centering
\small
\setlength{\tabcolsep}{5pt}
\renewcommand{\arraystretch}{1.08}
\caption{
Inter-annotator agreement for node-, edge-, and graph-level failure
labeling. Executions count eligible non-perfect executions receiving
secondary annotation; Judgments count their three level-wise decisions.
Agreement and unweighted Cohen's $\kappa$ are computed within each
benchmark by pooling the three level-wise judgments, using
pre-adjudication \emph{present}, \emph{absent}, and \emph{unknown} labels.
}
\label{tab:failure-label-agreement}
\begin{tabular}{lrrrr}
\toprule
Benchmark
& Executions
& Judgments
& Agreement
& Cohen's $\kappa$ \\
\midrule
Finance-Agent
& 62
& 186
& 92.5\%
& 0.85 \\
BrowseComp-Plus
& 135
& 405
& 90.1\%
& 0.81 \\
Plancraft
& 77
& 231
& 90.9\%
& 0.83 \\
WorkBench
& 105
& 315
& 90.5\%
& 0.83 \\
\bottomrule
\end{tabular}
\end{table}

\begingroup
\paragraph{Human adjudication results.}
The independent review summarized in
Table~\ref{tab:failure-label-agreement} comprises 1,137 level-wise
judgments across the four benchmarks, with 105 pre-adjudication
disagreements. Table~\ref{tab:human-label-adjudication} reports their
human adjudication outcomes. Human review supports the primary label
in 69 cases and the secondary label in 26 cases, assigns an alternative
label in six cases, and retains the unknown state in four unresolved
cases. These counts refer to level-wise judgments, not distinct
executions.
For the ten executions randomly sampled from records where both
annotators agree on all three levels, human review confirms
28 of the 30 level-wise judgments (93.3\%) and corrects two.
\endgroup

\begin{table}[htbp]
\centering
\caption{Human adjudication of pre-adjudication level-wise label
disagreements. Primary and Secondary denote the annotator whose label
is supported by human review; Neither denotes an alternative
human-assigned label. Unresolved cases retain the unknown state.}
\label{tab:human-label-adjudication}
\begingroup
\MashTableSetup
\begin{tabular*}{\linewidth}{@{\extracolsep{\fill}}lrrrrr@{}}
\toprule
Benchmark & Disagreements & Primary & Secondary & Neither & Unresolved \\
\midrule
Finance-Agent & 14 & 11 & 2 & 1 & 0 \\
BrowseComp-Plus & 40 & 26 & 10 & 3 & 1 \\
Plancraft & 21 & 13 & 5 & 1 & 2 \\
WorkBench & 30 & 19 & 9 & 1 & 1 \\
\midrule
Total & 105 & 69 & 26 & 6 & 4 \\
\bottomrule
\end{tabular*}
\endgroup
\end{table}

\subsection{Aggregation and Unknown Labels}
For method $m$, let $D_m$ be its non-perfect execution set and let
$y_{i,\ell}$ be the final label of execution $i$ at level $\ell$. The
confirmed issue count and its within-population percentage are
\[
  C_{m,\ell}
  =
  \sum_{i\in D_m}
  \mathbf{1}[y_{i,\ell}=\mathrm{present}],
  \qquad
  P_{m,\ell}
  =
  100\,\frac{C_{m,\ell}}{|D_m|}.
\]
Repeated occurrences at one level within an execution count once. The
coordination union is computed by execution identity:
\[
  C_{m,E\cup G}
  =
  \sum_{i\in D_m}
  \mathbf{1}
  [y_{i,E}=\mathrm{present}
  \ \lor\
  y_{i,G}=\mathrm{present}].
\]

\appendixneedspace{0.30\textheight}
\paragraph{Unknown-label accounting.}
Unknown labels remain in the denominator. For each method, benchmark,
and level, the three states exhaust the non-perfect execution set:
\[
  N^{\mathrm{present}}_{m,\ell}
  +
  N^{\mathrm{absent}}_{m,\ell}
  +
  N^{\mathrm{unknown}}_{m,\ell}
  =
  |D_m|.
\]
Executions with no present label are separated into those with three absent
labels and those containing at least one unknown label.

Table~\ref{tab:failure-label-unknown} reports unknown labels separately for
each method, benchmark, and level.

\begin{table}[H]
\centering
\small
\setlength{\tabcolsep}{5pt}
\renewcommand{\arraystretch}{1.08}
\caption{
Unknown labels in the execution-level failure analysis.
Each row identifies one benchmark--method population. Node, Edge, and
Graph count unknown labels at the corresponding level; these counts
need not refer to distinct executions. Unknown labels remain in the
denominator of within-method percentages.
}
\label{tab:failure-label-unknown}
\begin{tabular}{@{}llrrrr@{}}
\toprule
Benchmark & Method & Non-perfect & \multicolumn{3}{c}{Unknown labels} \\
\cmidrule(lr){4-6}
& & executions & Node & Edge & Graph \\
\midrule
Finance-Agent & MAS-Centralized & 117 & 4 & 2 & 2 \\
Finance-Agent & Meta-Harness & 101 & 3 & 2 & 1 \\
Finance-Agent & MASH & 91 & 3 & 1 & 1 \\
\midrule
BrowseComp-Plus & MAS-Centralized & 240 & 8 & 10 & 7 \\
BrowseComp-Plus & Meta-Harness & 225 & 7 & 9 & 7 \\
BrowseComp-Plus & MASH & 213 & 7 & 9 & 6 \\
\midrule
Plancraft & MAS-Centralized & 141 & 3 & 5 & 4 \\
Plancraft & Meta-Harness & 131 & 3 & 4 & 4 \\
Plancraft & MASH & 115 & 3 & 4 & 3 \\
\midrule
WorkBench & MAS-Centralized & 195 & 6 & 7 & 6 \\
WorkBench & Meta-Harness & 173 & 5 & 6 & 5 \\
WorkBench & MASH & 156 & 5 & 6 & 5 \\
\bottomrule
\end{tabular}
\end{table}

\appendixneedspace{0.18\textheight}
\section{Additional Related Work}
\label{app:harness-related-work}

\paragraph{Prompt, context, memory, and tool optimization.}
Prompt and context methods refine the information presented to the
agent: GEPA updates prompts through trajectory-based
reflection~\citep{agrawal2026gepa}, ACE accumulates strategies through
incremental context updates~\citep{zhang2025ace}, and ACON optimizes
context-compression guidelines from failure evidence~\citep{kang2026acon}.
Other methods change supporting components: MemEvolve modifies memory
implementation code~\citep{zhang2026memevolve}, while RewardHarness
evolves external tools and skills for a frozen image-preference
evaluator~\citep{zhang2026rewardharness}.

\paragraph{Executable harness optimization.}
Executable harness methods revise code and interfaces using execution
evidence. Meta-Harness draws on historical code and
traces~\citep{lee2026metaharness}; AHE verifies edits using evidence and
predicted effects~\citep{lin2026ahe}; Life-Harness intervenes across the
agent--environment lifecycle~\citep{xu2026lifeharness}. HarnessX and Self-Harness link
failure evidence to modifications and acceptance
checks~\citep{chen2026harnessx,zhang2026selfharness}, while AutoSaddler
uses execution-trace diagnosis for targeted updates~\citep{autosaddler2026}.

\paragraph{Evaluation efficiency and continual adaptation.}
HarnessLens selects informative validation tasks~\citep{harnesslens2026};
HarnessBank uses gated verification~\citep{harnessbank2026}; ModularRSI
applies successive program, edit, and execution checks~\citep{wu2026modularrsi}.
EvoTest, Continual Harness, and Evo-Harness extend adaptation across
repeated episodes, interactions, or tasks~\citep{he2026evotest,continualharness2026,evoharness2026}.
These settings differ in when feedback is available and whether
adaptation continues during evaluation, motivating explicit feedback
and inference budgets together with independent held-out
assessment~\citep{rethinkingharness2026,harnessoptbench2026}.

\end{document}